\documentclass[10pt]{article} 
\usepackage[accepted]{tmlr}

\usepackage{amsmath,amsfonts,bm}

\def\eqref#1{equation~\ref{#1}}

\def\1{\bm{1}}

\DeclareMathAlphabet{\mathsfit}{\encodingdefault}{\sfdefault}{m}{sl}
\SetMathAlphabet{\mathsfit}{bold}{\encodingdefault}{\sfdefault}{bx}{n}

\usepackage{hyperref}
\usepackage{url}
\usepackage{graphicx} 
\usepackage{wrapfig}
\usepackage{amsmath}
\usepackage{booktabs}
\usepackage{subcaption}
\usepackage{multirow}
\usepackage{marvosym}
\usepackage{amssymb}
\usepackage{amsbsy}
\usepackage{pifont}
\usepackage{caption}

\newcommand{\cmark}{\ding{51}}
\newcommand{\xmark}{\ding{55}}

\newcommand{\rot}[1]{\rotatebox{90}{#1}}

\usepackage{colortbl}

\definecolor{bgreen}{RGB}{0,170,0}
\definecolor{bred}{RGB}{220,0,0}
\definecolor{mydarkblue}{RGB}{0,0,255}
\definecolor{Gray}{gray}{0.93}
\definecolor{skyblue}{rgb}{0.925,0.957,1}
\definecolor{PineGreen}{RGB}{1, 121, 111}
\definecolor{RedBrick}{RGB}{77,0,38}

\newcommand{\ours}{\cellcolor{skyblue}}
\newcommand{\improve}[1]{\textcolor{PineGreen}{$\uparrow #1$}}

\title{Domain Generalizable Adaptation of 3D Vision-Language Models via Regularized Fine-Tuning}

\author{\name Sneha Paul \email sneha.paul@mail.concordia.ca \\
      \addr Concordia University, Canada
      \AND
      \name Zachary Patterson \email zachary.patterson@concordia.ca \\
      \addr Concordia University, Canada
      \AND
      \name Nizar Bouguila \email nizar.bouguila@concordia.ca \\
      \addr Concordia University, Canada}

\def\month{06}  
\def\year{2026} 
\def\openreview{\url{https://openreview.net/forum?id=453uT7O7wc}} 

\begin{document}

\maketitle

\begin{abstract}
Domain adaptation remains a central challenge in 3D vision, especially for multimodal foundation models that align 3D point clouds with visual and textual data. While these models demonstrate strong general capabilities, adapting them to downstream domains with limited data often leads to overfitting and catastrophic forgetting. To address this, we introduce ReFine3D, a regularized fine-tuning framework designed for domain-generalizable tuning of 3D large multimodal models (LMMs). ReFine3D combines selective layer tuning with two targeted regularization strategies: multi-view consistency across augmented point clouds and text diversity through synonym-based prompts generated by large language models. Additionally, we incorporate point-rendered vision supervision and a test-time augmentation mechanism with confidence-based aggregation to further enhance robustness. Extensive experiments across different 3D domain generalization benchmarks show that ReFine3D improves base-to-novel class generalization by 1.36\%, cross-dataset transfer by 2.43\%, robustness to corruption by 1.80\%, and few-shot accuracy by up to 3.11\%---outperforming prior state-of-the-art methods with minimal added computational overhead.
\end{abstract}

\section{Introduction}
\label{sec:intro}
Point clouds are fundamental to modern 3D perception systems, powering applications from autonomous driving to augmented reality by capturing fine-grained spatial and geometric details of real-world environments. A fundamental task in these systems is recognizing 3D objects from point cloud data. Unlike structured modalities such as images or voxels, point clouds are unordered, sparse, and irregular --- making them both rich in information and challenging to process. To address these challenges and build robust point cloud foundation models, recent work has adopted multimodal learning strategies (inspired by advances in vision-language models, such as CLIP~\citep{clip}) by aligning 3D representations with corresponding visual and textual data~\citep{pointclip, pointclip2, zhang2023learning, ulip, ulip2, paul2026point}. While these approaches show promise in capturing transferable features, adapting them to specific downstream tasks remains difficult, particularly when labelled data is scarce and has a strong distribution shift. Fine-tuning these large multimodal models (LMMs) on limited downstream data can lead to overfitting and a loss of the broad, transferable knowledge acquired during pre-training --- a phenomenon known as \textit{catastrophic forgetting}~\citep{qi2023fine}. This degrades the model's ability to generalize to unseen domains or corrupted data, making naive fine-tuning not only ineffective but potentially harmful in real-world, domain-shifted scenarios.
 
To enable efficient fine-tuning of LMMs with limited labelled data, parameter-efficient fine-tuning (PEFT) strategies such as adapter tuning~\citep{gao2024clip, song2023meta, paul2026adapter} and prompt tuning~\citep{promptsrc, promptkd, coop} have been proposed. Following the success of such techniques in vision-language models, recent work in the 3D domain (e.g. PointPRC \citep{pointprc}) has adopted similar PEFT strategies for fine-tuning pre-trained 3D LMMs. However, we identify several key limitations of existing PEFT approaches in the context of 3D large multi-modal models. 
\textbf{First}, unlike image or language encoders, existing 3D encoders typically have significantly fewer parameters. For example, PointBERT \citep{pointbert}, a representative 3D encoder, contains 22.8 million parameters, compared to the 86 million parameter image encoder in the base version of CLIP \citep{clip}. This lower capacity makes 3D encoders inherently less prone to overfitting under limited data. 
\textbf{Second}, point clouds differ fundamentally from images: they are unordered, unstructured, and sparse, lacking the regular grid-like organization of 2D data~\citep{guo2020deep}. Consequently, methods that are effective for the image domain may not necessarily perform well for the point cloud. Nonetheless, most existing PEFT strategies for point clouds naively adopt prompt tuning techniques developed for images, overlooking the distinct properties of 3D data.
\textbf{Third}, state-of-the-art (SOTA) 3D prompt tuning methods such as PointPRC \citep{pointprc} fine-tune the multimodal encoder by aligning point cloud features with textual representations, while discarding the image encoder during downstream adaptation. This leads to suboptimal learning, as the image encoder, pre-trained on large-scale image-text pairs, contains rich semantic knowledge that could be especially beneficial in low-resource 3D scenarios.

\begin{figure*}
    \centering
    \includegraphics[width=0.9\linewidth]{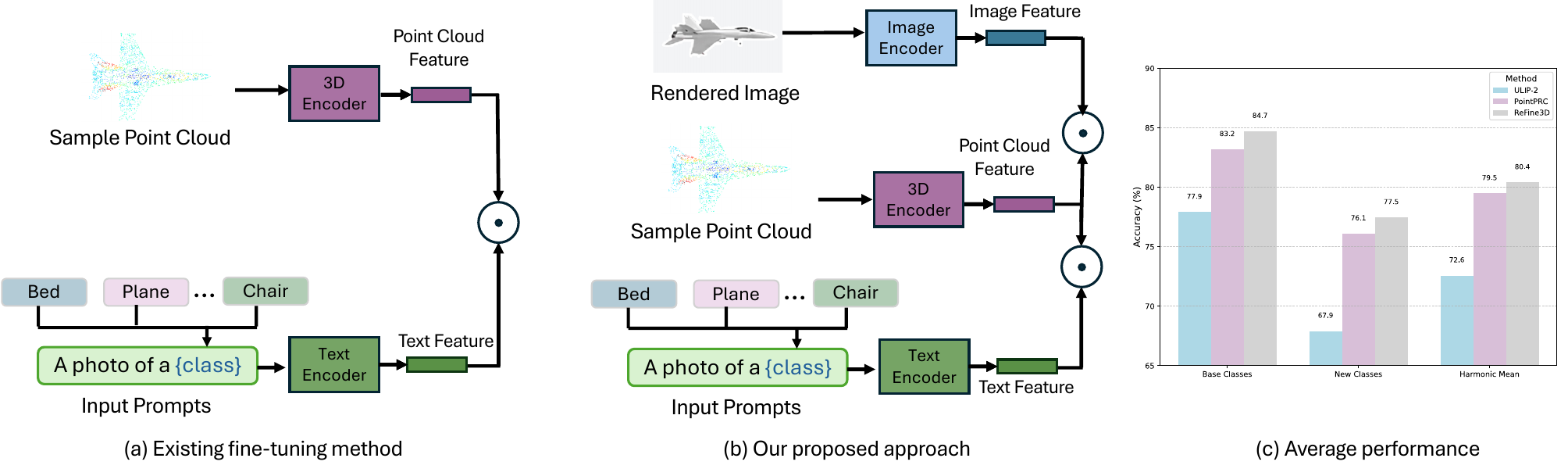}%
    \caption{Unlike existing methods that discard the pre-trained vision encoder of 3D VLMs, shown in (a), our proposed fine-tuning framework utilizes the pre-trained visual priors by aligning the 3D, vision and text features in a shared representation space (presented in (b)). 
    This strategy, along with our other proposed tuning strategies, improves the state-of-the-art in 3D domain generalizations (shown in (c)). 
    }
    \label{fig:banner}
\vspace{-20pt}
\end{figure*}

Motivated by these observations, in this work, we propose \textbf{Re}gularized \textbf{Fine}-tuning framework (\textbf{ReFine3D}) for tuning 3D multimodal foundation models. ReFine3D enables effective adaptation to downstream tasks while preserving the generalizable knowledge of the pre-trained models. Rather than relying on prompt tuning, ReFine3D introduces a layer-selective fine-tuning approach that strategically fine-tunes a subset of layers within the pre-trained encoder. 
Furthermore, to prevent overfitting due to additional layer tuning and promote generalization under low-data scenarios during fine-tuning, we incorporate two regularization strategies. The first strategy is multi-modal consistency regularization. Instead of enforcing consistency between the point cloud embedding and the class embedding (like PointPRC), we apply multiple augmentations to the input point cloud and ensure consistency across the embeddings of all augmented samples. This approach mitigates memorization and reduces overfitting to specific input instances. 
The second regularization is a text diversity regularization. Rather than generating text embedding directly from the class name (like PointPRC) using a hand-crafted prompt template, we first derive multiple synonyms for the class using WordNet \citep{miller1995wordnet}. These synonyms are then used by a large pre-trained language model (LLM) to generate diverse descriptive sentences. We enforce consistency between the point cloud embedding and the embeddings of these sentences, promoting robustness and reducing overfitting. Finally, to retain and leverage the pre-trained vision encoder's knowledge during fine-tuning, we propose a point-rendered vision supervision method. This approach renders images from input point clouds and aligns their representations with both 3D and text embeddings, improving multi-modal consistency and performance.
During inference, ReFine3D introduces a test-time augmentation mechanism with confidence-based aggregation for improving the performance. Specifically, we generate multiple augmentations of the input point cloud and apply the text diversification technique to create diverse class descriptions of the text classes. We then select the top-$H$ embeddings with the highest confidence (based on maximum softmax probability) and aggregate their predictions via majority voting. While the components in ReFine3D are not entirely novel individually, the novelty of this work lies in combining all the components into a novel fine-tuning framework for 3D LMMs to solve a very specific problem: effective domain-generalizable adaptation of 3D multimodal foundation models under limited data, while mitigating overfitting and preserving pre-trained generalization capabilities.

To evaluate our proposed framework, we conduct comprehensive experiments following the standard 3D domain generalization (3D-DG) protocol \citep{pointprc}, covering base-to-new generalization on five datasets (ModelNet40 \citep{modelnet}, ShapeNetCoreV2 \citep{shapenet}, and three ScanObjectNN variants \citep{scanobject}), cross-dataset generalization under four settings (OOD generalization, data corruption, domain adaptation, and sim-to-real transfer), and few-shot learning with 1, 2, 4, 8, and 16-shot settings. Our results show consistent gains across all benchmarks. Notably, ReFine3D achieves a 1.53\% improvement on base classes and 1.36\% on novel classes in the base-to-new setting, with a 0.92\% average gain in harmonic mean across five datasets. For cross-dataset generalization, ReFine3D yields a 2.43\% average improvement across five target domains. Under various types of data corruption, it outperforms the SOTA by 1.80\% on average. In the few-shot setting, our method demonstrates strong generalization, achieving a 3.11\% gain even with only 1-shot supervision. Furthermore, to validate the significance of each tuning strategy, we perform extensive ablation and sensitivity analysis, and provide computational cost evaluation. 
Overall, we make the following contributions in this paper: 
\begin{itemize}
    \item To address the challenges of existing 3D fine-tuning literature, we propose ReFine3D, a regularized fine-tuning framework that selectively fine-tunes specific layers of the pre-trained 3D VLMs for domain generalizable adaptation.
    \item We introduce two regularization strategies: augmentation-based consistency and text synonymization, to prevent overfitting and enhance robustness, especially under limited labelled data.
    \item We propose point-rendered vision supervision, which explicitly leverages the frozen CLIP image encoder's visual priors from tri-modal pre-trained models, discarded by other existing fine-tuning methods, enabling better cross-modal representation learning.
    \item Finally, we develop a test-time augmentation mechanism with confidence-based aggregation that aggregates predictions across multiple augmented views and textual variations during inference to boost performance.
\end{itemize}

\section{Related Work}
\label{sec:related_work}

\subsection{Multi-Modal Vision-Language Models in 3D point clouds}
The integration of vision-language models (VLMs) into 3D point cloud analysis has revolutionized the field, enabling models to leverage both visual and textual modalities for improved generalization and open-vocabulary understanding. Pioneering works like PointCLIP \citep{pointclip}, pointclip2 \citep{pointclip2}, CLIP2Point \citep{clip2point}, ULIP \citep{ulip}, ULIP2 \citep{ulip2} extended the success of CLIP \citep{clip} to 3D by projecting point clouds into a shared embedding space with text. By aligning point cloud features with semantically rich text embeddings, these models achieved remarkable zero-shot generalization, allowing them to recognize objects from unseen categories without task-specific fine-tuning. 
Despite their advancements, these models share several limitations. First, they rely on full fine-tuning for downstream tasks, which is computationally expensive and risks overfitting to specific datasets \citep{wortsman2022robust}. Second, while they excel in zero-shot settings, their performance degrades significantly under domain shifts, such as variations in sensor data, environmental conditions, or object appearances \citep{zhou2022domain}. For instance, models trained on synthetic datasets like ShapeNet struggle to generalize to real-world scans from ScanObjectNN or corrupted data from ModelNet-C \citep{modelnetC}. Finally, these models often lack in explicitly enforcing cross-modal consistency during fine-tuning, leading to misaligned representations that hinder generalization.
These limitations highlight the need for more robust and efficient adaptation strategies, particularly in scenarios where labelled data is scarce or domain shifts are prevalent. 

\vspace{-5pt}
\subsection{3D Domain Generalization (3D-DG)}
\vspace{-5pt}
3D domain generalization (3D-DG) aims to develop models that perform robustly across unseen domains without requiring additional fine-tuning. This is particularly challenging in real-world applications, where 3D point cloud data can vary significantly due to factors like sensor noise, occlusions, or geometric transformations. Early approaches to 3D-DG, such as PointDAN \citep{pointdan} and MetaSets \citep{metasets}, focused on domain adaptation and meta-learning to improve generalization. PointDAN introduced a multi-scale feature alignment strategy to bridge the gap between source and target domains, while MetaSets employed meta-learning on transformed point sets to handle sim-to-real geometry shifts. However, these methods were limited to small-scale datasets (e.g., ModelNet with fewer than 10,000 samples) and architectures (e.g., PointNet with 1.2M parameters), struggling to scale to the complexity of modern 3D tasks.
More recent works have sought to address these limitations by leveraging large-scale pre-trained models and multi-modal learning. For instance, PDG \citep{pdg} proposed a part-level domain generalization framework, decomposing 3D objects into shared part spaces to reduce domain gaps. While effective, PDG still relies on small datasets and lacks the scalability of modern foundation models.
\vspace{-5pt}

\subsection{Parameter-Efficient Fine-Tuning (PEFT) for Domain Generalization (DG)}
Recent literature \citep{pointPEFT, ppt, idpt, dept} has explored parameter-efficient fine-tuning (PEFT) techniques as a solution to the above-mentioned challenges. By introducing a small number of learnable parameters, PEFT enables efficient adaptation of pre-trained foundation models to downstream tasks while preserving their generalization capability. For instance, PPT \citep{ppt} and PointPEFT \citep{pointPEFT} demonstrated that prompt tuning can significantly improve task-specific performance without extensive retraining. These methods align point cloud features with semantically rich text embeddings, leveraging the strengths of multi-modal vision-language models like PointCLIP \citep{pointclip} and ULIP \citep{ulip}.

Methods such as PointPEFT, IDPT, and related approaches are designed for point-cloud-only backbones trained on task-specific, in-distribution datasets (e.g., single-object classification). In contrast, our work focuses on adapting large-scale 3D vision-language models pre-trained on multimodal data, for which the objective is to preserve and improve cross-domain generalization rather than optimize performance on a single supervised task. Due to this fundamental difference in model paradigm (unimodal task-specific training vs. multimodal pre-trained vision-language models) and evaluation setting (in-distribution task performance vs. base-to-new generalization), a direct empirical comparison is not appropriate.
However, existing PEFT-based approaches for 3D-DG face several limitations. First, many methods rely on multi-modal prompt tuning, which fine-tunes both point cloud and text encoders, introducing unnecessary complexity and increasing the risk of overfitting, particularly when text encoders are adapted to narrow datasets. 
Second, the evaluation of 3D-DG methods is hindered by the lack of diverse and robust benchmarks. While datasets like ModelNet-C \citep{modelnetC} and PointDA \citep{pointdan} provide some evaluation scenarios, they often fail to capture the full spectrum of real-world domain shifts, such as cross-dataset generalization or few-shot adaptation. Furthermore, the integration of semantic diversity into prompt learning remains underexplored. Current methods \citep{pointprc} primarily rely on LLM-generated or hand-crafted text descriptions, which may lack the structured semantic richness offered by external knowledge bases like WordNet \citep{miller1995wordnet}.

Our work addresses these limitations by introducing a systematic fine-tuning framework for 3D multi-modal foundation models that balances task-specific adaptation with generalization. By selectively updating 3D encoder layers and incorporating consistency-based regularization across point cloud and text modalities, and utilizing the pre-trained image encoder, our work enhances robustness to domain shifts while preserving the rich priors of pre-trained vision-language models.

\section{Methodology}
\label{sec:method}

\subsection{Preliminaries}
We begin by outlining the foundational components of our approach, building on vision-language and 3D multi-modal models. The Contrastive Language-Image Pre-training (CLIP) model~\citep{clip} learns aligned representations of images and text by maximizing the similarity between paired image-text embeddings while minimizing similarity for unpaired samples. Given an image embedding $\mathbf{z}_I \in \mathbb{R}^d$ and a text embedding $\mathbf{z}_T \in \mathbb{R}^d$, CLIP performs zero-shot classification by computing the similarity $\text{sim}(\mathbf{z}_I, \mathbf{z}_T) = \frac{\mathbf{z}_I \cdot \mathbf{z}_T}{\|\mathbf{z}_I\| \|\mathbf{z}_T\|}$ and selecting the class with the highest similarity to the input image embedding.

The Unified Language-augmented Point Cloud (ULIP) model~\citep{ulip} extends this framework to tri-modal alignment between point clouds, images, and text. ULIP processes all three modalities through separate encoders: (1) a point cloud encoder $f_P$ generating $\mathbf{z}_P \in \mathbb{R}^d$ from input point clouds $\mathbf{x} \in \mathbb{R}^{N \times 3}$, (2) an image encoder $f_I$ producing $\mathbf{z}_I \in \mathbb{R}^d$ from rendered 2D views of a point cloud, and (3) a text encoder $f_T$ yielding $\mathbf{z}_T \in \mathbb{R}^d$ from language descriptions. The model learns a unified embedding space through a symmetric contrastive loss that enforces similarity between all positive triplets $(\mathbf{z}_P, \mathbf{z}_I, \mathbf{z}_T)$ while pushing apart negative pairs. However, while ULIP's pre-trained knowledge captures generalizable 3D geometric and semantic features, it often fails to adapt to new domains or tasks with subtle geometric cues, necessitating fine-tuning for downstream tasks.

For fine-tuning, existing literature \citep{pointprc} adopts a supervised approach using the cross-entropy (CE) loss. Given a point cloud sample $\mathbf{x} \in \mathbb{R}^{N \times 3}$ (where $N$ is the number of points) and its class label $y \in \{1, \dots, C\}$, the method computes the point cloud embedding $\mathbf{z}_P = f_P(\mathbf{x})$ using the point cloud encoder $f_P$. The class prediction is obtained by comparing $\mathbf{z}_P$ with text embeddings $\{\mathbf{z}_T^c\}_{c=1}^C$ for each class, and the model is optimized using:
\begin{equation}
\label{eq:1}
\mathcal{L}_{\text{CE}} = -\log 
\frac{\exp(\text{sim}(\mathbf{z}_P, \mathbf{z}_T^{y}) / \tau)}
{\sum_{c=1}^C \exp(\text{sim}(\mathbf{z}_P, \mathbf{z}_T^{c}) / \tau)},
\end{equation}

\begin{wrapfigure}{r!}{0.5\textwidth}
\vspace{-20pt}
\begin{center}
    \centering
    \includegraphics[width=1.0\linewidth]{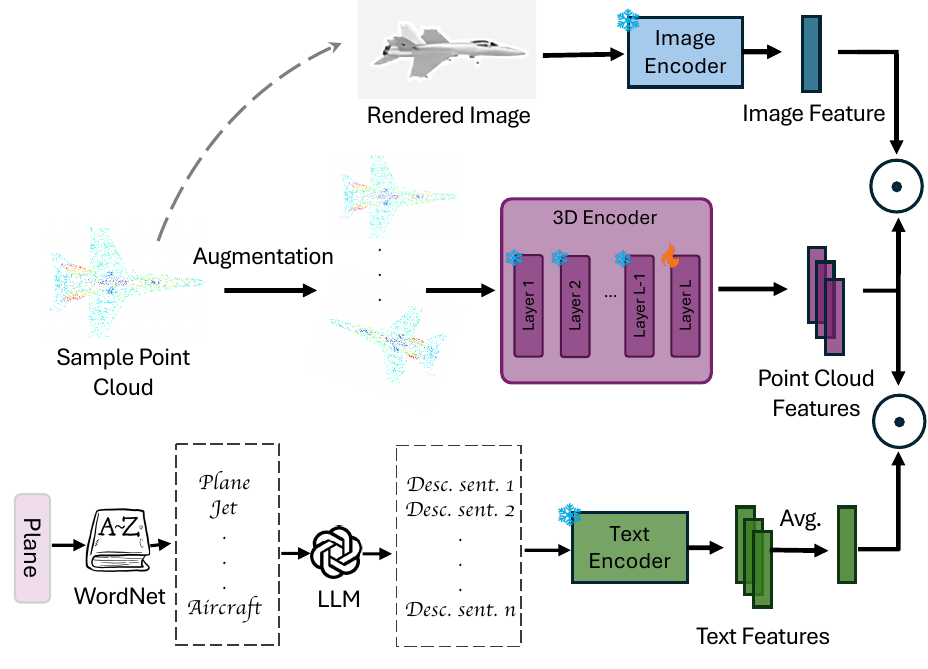}
    \caption{Our proposed Regularized Fine-tuning Framework, ReFine3D, for tuning 3D VLMs that selectively updates encoder layers while utilizing the pre-trained vision encoder’s knowledge. It introduces various tuning strategies, such as layer-selective fine-tuning, augmentation- and synonym-based regularization during training, and test-time augmentation to improve task-specific learning without forgetting pre-trained task-agnostic knowledge.}
    \label{fig:method}
\end{center}
\vspace{-40pt}
\end{wrapfigure}
where $\tau$ is a temperature parameter. 
This multi-modal fine-tuning approach preserves ULIP's cross-modal alignment while adapting to downstream tasks through discriminative learning.
Despite the success of fine-tuning approaches in 3D vision-language models, we identify key limitations in existing methods, including solely relying on PEFT for tuning, discarding the pre-trained image encoder, and simply following image literature. In the following section, we discuss our proposed method, ReFine3D, motivated by these limitations.

\subsection{ReFine3D}

We propose \textbf{ReFine3D}, a novel framework for fine-tuning 3D LMMs for domain-generalizable adaptation while preserving pre-trained knowledge. 
Our approach is designed for 3D vision-language models that follow CLIP-style symmetric tri-modal contrastive pre-training, where point clouds ($\mathbf{z}_P$), images ($\mathbf{z}_I$), and text ($\mathbf{z}_T$) are mutually aligned in a shared embedding space through symmetric contrastive objectives during pre-training~\citep{ulip,paul2023crossmoco, paul2024improving}. This tri-modal architecture enables ReFine3D to leverage complementary supervision from both modalities during fine-tuning: the frozen image encoder $f_I$ provides rich visual priors learned from large-scale image-text pairs, while the frozen text encoder $f_T$ enables semantic guidance through diversified prompts. Unlike recent alternatives that use direct 3D-to-text alignment (e.g., Uni3D~\citep{uni3d}) or asymmetric multi-task learning (e.g., OpenShape~\citep{openshape}), our framework requires explicit tri-modal alignment where all three encoders are available and the image/text encoders remain frozen to preserve pre-trained knowledge.

\begin{table*}[t]
 \vspace{-30pt}
    \centering
    \small
    \caption{\textbf{Complete hyperparameter configuration for ReFine3D.} All experiments use these settings unless otherwise specified.}
    \label{tab:hyperparameters}
    \begin{tabular}{l l l}
        \toprule
        \textbf{Component} & \textbf{Parameter} & \textbf{Value} \\
        \midrule
        \multirow{6}{*}{\textbf{Architecture}} 
        & Backbone & ULIP-2 with PointBERT encoder \\
        & 3D encoder layers ($L$) & 12 \\
        & Frozen layers ($L_f$) & 11 (layers 1--11 frozen) \\
        & Trainable layers & 1 (layer 12 fine-tuned) \\
        & Image encoder & CLIP ViT-B/16 (frozen) \\
        & Text encoder & CLIP text encoder (frozen) \\
        \midrule
        \multirow{5}{*}{\textbf{Training Augmentation}} 
        & Number of augmentations ($K$) & 16 \\
        & Rotation range & [0°, 360°] on z-axis \\
        & Scale range & [0.8, 1.2] \\
        & Jitter std. deviation & 0.01 \\
          
        \midrule
        \multirow{4}{*}{\textbf{Text Diversity}} 
        & LLM for prompt generation & Qwen-2.5-7B-Instruct \\
        & Number of synonyms (WordNet) & 7 per class \\
        & Prompts per class ($Q$) & 7 \\
        & Prompt template & ``A 3D point cloud of \{synonym\}'' \\
        \midrule
        \multirow{6}{*}{\textbf{Rendering Configuration}} 
        & Renderer & Blender with BlenderProc \\
        & Views per object & 12 \\
        & Camera distance & 1.8 meters from centroid \\
        & Azimuthal angles & 0°--360° (30° increments) \\
        & Elevation angles & 0°, 15°, 30° \\
        & Image resolution & 512$\times$512 pixels \\
        & Focal length & 35mm \\
        & Lighting & Directional (intensity 1.0) + Ambient (0.25) \\
        \midrule
        \multirow{5}{*}{\textbf{Loss \& Optimization}} 
        & Contrastive weight ($\alpha$) & 1.0 \\
        & Temperature ($\tau$) & 0.07 \\
        & Optimizer & SGD \\
        & Learning rate & 0.0025 \\
        & Momentum & 0.99 \\
        & Weight decay & 1e-5\\
        & LR schedule &  Cosine annealing  \\
        & Training epochs & 20 \\
        & Batch size & 32 \\
        \midrule
        \multirow{4}{*}{\textbf{Test-time Augmentation}} 
        & Test augmentations ($K_{test}$) & 5 \\
        & Text prompts per class ($Q$) & 7 \\
        & Top-$H$ selection & 3 \\
        & Confidence metric & Maximum softmax probability \\
        & Aggregation method & Majority voting \\
        \midrule
        \multirow{2}{*}{\textbf{Other}} 
        & Random seeds & 3 runs \\
        & Point cloud sampling & 10,000 points (ModelNet40, ShapeNet) \\
        \bottomrule
    \end{tabular}
\end{table*}
\vspace{30pt}

\begin{table*}[t]
    \small
   \centering
   \caption{\textbf{Performance across 5 benchmarks on Base-to-new class generalization on 3D Vision Language Models (VLMs).} We adopt the performance of existing 3D VLMs with prompt tuning from \citep{pointprc} and compare them with our proposed fine-tuning framework, ReFine3D. We consider ULIP and ULIP-2 as the pre-trained models since it is the most widely used models by existing fine-tuning methods. 
   Here, Base: base class accuracy (\%). New: new class accuracy (\%). HM: harmonic mean of base and new class accuracy (\%). Improvements are reported over existing SOTA, \textit{PointPRC}.
   }
   \begin{subtable}{0.30\linewidth}
      \caption{\textbf{Average over 5 datasets}}
      \setlength{\tabcolsep}{2pt}
      \centering
      \resizebox{1.0\textwidth}{!}{
      \begin{tabular}{l c c | c}
      \toprule
      Method & Base & New & HM \\
      \midrule
      P-CLIP~\citep{pointclip} & 75.66 & 23.45 & 35.80 \\
      P-CLIP2~\citep{pointclip2} & 74.11 & 37.84 & 50.10 \\ %
      \midrule
      ULIP~\citep{ulip} & 77.32 & 49.01 & 59.99 \\
      PointPRC~\citep{pointprc} & \textit{82.19} & \textit{61.93} & \textit{70.64} \\
      \rowcolor{skyblue}
      \textbf{ReFine3D} & \textbf{84.59} & \textbf{65.07} & \textbf{73.29} \\
      \midrule
      ULIP-2~\citep{ulip2} & 77.91 & 67.91 & 72.57 \\ %
      PointPRC~\citep{pointprc} & \textit{83.18} & \textit{76.10} & \textit{79.48} \\
      \rowcolor{skyblue}
      \textbf{ReFine3D} & \textbf{84.71} & \textbf{77.46} & \textbf{80.40} \\
      \bottomrule
      \end{tabular}
      }
      \label{tab:base2new_avg_five_datasets}
   \end{subtable}
   \quad
   \begin{subtable}{0.30\linewidth}
      \caption{ModelNet40}
            \setlength{\tabcolsep}{2pt}
      \centering
      \resizebox{1.0\textwidth}{!}{
      \begin{tabular}{l c c | c}
      \toprule
      Method & Base & New & HM \\
      \midrule
      P-CLIP~\citep{pointclip} & 93.23 & 20.22 & 33.23\\
      P-CLIP2~\citep{pointclip2} & 93.98 & 45.21 & 61.05\\
      \midrule
      ULIP~\citep{ulip} & 92.80 & 50.07 & 65.05 \\
      PointPRC~\citep{pointprc}& 95.03 & 55.27 & 69.89 \\
       \rowcolor{skyblue}\textbf{ReFine3D} & \textbf{97.17} & \textbf{58.30} & \textbf{72.88} \\
      \midrule
      ULIP-2~\citep{ulip2} & 91.77 & 56.47 & 69.92 \\ %
      PointPRC~\citep{pointprc} & \textit{95.30} & \textit{64.83} & \textit{77.17}\\
      \rowcolor{skyblue}\textbf{ReFine3D} & \textbf{95.87} & \textbf{66.23} & \textbf{78.34} \\
      \bottomrule
      \end{tabular}
      }
      \label{tab:base2new_mn40}
   \end{subtable}
   \quad
   \begin{subtable}{0.30\linewidth}
      \caption{S-PB\_T50\_RS}
      \setlength{\tabcolsep}{2pt}
      \centering
        \resizebox{1.0\textwidth}{!}{
        \begin{tabular}{l c c | c}
      \toprule
      Method & Base & New & HM \\
      \midrule
      P-CLIP~\citep{pointclip} & 61.25 & 19.87 & 30.01\\
      P-CLIP2~\citep{pointclip2} & 56.84 & 29.92 & 39.20\\
      \midrule
      ULIP~\citep{ulip} & 56.73 & 25.80 & 35.47 \\
      PointPRC~\citep{pointprc}& \textit{64.20} & \textit{49.17} & \textit{55.69} \\
     \rowcolor{skyblue} \textbf{ReFine3D} & \textbf{66.76} & \textbf{52.89} & \textbf{59.02} \\
      \midrule
      ULIP-2~\citep{ulip2} & 66.40 & 66.47 & 66.43 \\
      PointPRC~\citep{pointprc} & \textit{73.67} & \textit{74.27} & \textit{73.97}\\
      \rowcolor{skyblue} \textbf{ReFine3D} & \textbf{76.00} & \textbf{75.80} & \textbf{75.90} \\
      \bottomrule
      \end{tabular}
      }
      \label{tab:base2new_so_pb_t50_rs}
   \end{subtable}
   \begin{subtable}{0.30\linewidth}
      \vspace{9pt}
      \caption{S-OBJ\_BG}
            \setlength{\tabcolsep}{2pt}
      \centering
            \resizebox{1.0\textwidth}{!}{
      \begin{tabular}{l c c | c}
      \toprule
      Method & Base & New & HM \\
      \midrule
      P-CLIP~\citep{pointclip} & 72.82 & 23.00 & 34.96 \\
      P-CLIP2~\citep{pointclip2} & 70.07 & 35.08 & 46.75\\
      \midrule
      ULIP~\citep{ulip} & 73.20 & 47.17 & 57.37 \\
      PointPRC~\citep{pointprc} & \textit{79.47} & \textit{55.20} & \textit{65.15} \\
      \rowcolor{skyblue} \textbf{ReFine3D} & \textbf{82.05} & \textbf{59.77} & \textbf{69.51} \\
      \midrule
      ULIP-2~\citep{ulip2} & 77.00 & 83.27 & 80.01 \\
      PointPRC~\citep{pointprc} & \textit{80.10} & \textit{88.93} & \textit{84.28}\\
      \rowcolor{skyblue} \textbf{ReFine3D} & \textbf{82.35} & \textbf{90.12} & \textbf{86.06} \\
      \bottomrule
      \end{tabular}
      }
      \label{tab:base2new_so_obj_bg}
   \end{subtable}
   \quad
   \begin{subtable}{0.30\linewidth}
      \vspace{9pt}
      \caption{S-OBJ\_ONLY}
    \setlength{\tabcolsep}{2pt}
          \resizebox{1.0\textwidth}{!}{
      \centering
      \begin{tabular}{l c c | c}
      \toprule
      Method & Base & New & HM \\
      \midrule
      P-CLIP~\citep{pointclip} & 76.23 & 20.23 & 31.97\\
      P-CLIP2~\citep{pointclip2} & 71.40 & 44.39 & 54.74\\
      \midrule
      ULIP~\citep{ulip} & 74.13 & 50.80 & 60.29 \\
      PointPRC~\citep{pointprc} & \textit{79.23} & \textit{65.93} & \textit{71.97} \\
      \rowcolor{skyblue} \rowcolor{skyblue} \textbf{ReFine3D} & \textbf{82.65} & \textbf{68.26} & \textbf{74.77} \\
      \midrule
      ULIP-2~\citep{ulip2} & 78.60 & 76.27 & 77.42 \\
      PointPRC~\citep{pointprc} & \textit{83.60} & \textit{81.10} & \textit{82.33}\\
      \rowcolor{skyblue} \rowcolor{skyblue} \textbf{ReFine3D} & \textbf{84.05} & \textbf{82.97} & \textbf{83.51} \\
      \bottomrule
      \end{tabular}
      }
      \label{tab:base2new_so_obj_only}
   \end{subtable}
   \quad
   \begin{subtable}{0.30\linewidth}
      \vspace{9pt}
      \caption{ShapeNetCoreV2}
      \setlength{\tabcolsep}{2pt}
    \resizebox{1.0\textwidth}{!}{
      \centering
      \begin{tabular}{l c c | c}
      \toprule
      Method & Base & New & HM \\
      \midrule
      P-CLIP~\citep{pointclip} & 74.78 & 33.92 & 46.61\\
      P-CLIP2~\citep{pointclip2} & 78.27 & 34.58 & 47.97\\
      \midrule
      ULIP~\citep{ulip} & 89.73 & 71.20 & 79.40 \\
      PointPRC~\citep{pointprc} & \textit{93.03} & \textit{84.10} & \textit{88.34} \\
      \rowcolor{skyblue} \textbf{ReFine3D} & \textbf{94.82} & \textbf{86.12} & \textbf{90.26} \\
      \midrule
      ULIP-2~\citep{ulip2} & 75.80 & 57.07 & 65.38 \\ %
      PointPRC~\citep{pointprc} & \textit{83.23} & \textit{71.37} & \textit{76.85}\\
      \rowcolor{skyblue} \textbf{ReFine3D} & \textbf{85.31} & \textbf{72.62} & \textbf{78.46} \\
      \bottomrule
      \end{tabular}
      }
      \label{tab:base2new_snv2}
   \end{subtable}
   \label{tab:base2new}
\end{table*}

Unlike existing prompt-tuning approaches, ReFine3D selectively fine-tunes specific layers of the pre-trained point cloud encoder to enable downstream adaptation. Consider a transformer-based point cloud encoder $f_P$ with $L$ layers, denoted as $\{l_1, l_2, \dots, l_L\}$. We freeze the first $L_f$ layers ($l_1$ to $l_{L_f}$) to retain pre-trained knowledge and fine-tune the last $L - L_f$ layers ($l_{L_f+1}$ to $l_L$) to capture task-specific features. Let $\theta_f$ and $\theta_t$ denote the parameters of the frozen and trainable layers, respectively. The point cloud embedding is computed as:
\begin{equation}
\mathbf{z}_P = f_P(\mathbf{x}; \theta_f, \theta_t).
\end{equation}
This approach balances adaptation and generalization, as the early layers encode low-level geometric features that are broadly transferable, while the later layers capture high-level semantics that benefit from task-specific tuning. 
 
Unlike convolutional architectures that may encode certain geometric priors, our point cloud encoder (PointBERT, detailed in Section 3.3) is a transformer-based architecture that operates directly on raw 3D coordinates without built-in invariance to geometric transformations such as rotation, scaling, or translation. Therefore, we explicitly enforce robustness to these transformations through data augmentation during training.
To prevent overfitting during fine-tuning, we introduce two regularization constraints.
Rather than enforcing consistency between a single point cloud embedding and its class embedding, we apply augmentation-based consistency regularization, where $K$ geometric transformations are applied to the input point cloud $\mathbf{x}$, generating augmented samples $\{\mathbf{x}_k\}_{k=1}^K$. Common augmentations include random rotation, scale, and jitter. Each augmented sample is passed through the encoder to obtain embeddings $\{\mathbf{z}_P^k = f_P(\mathbf{x}_k)\}_{k=1}^K$. We compute the CE loss for each augmented sample:
\begin{equation}
\mathcal{L}_{\text{CE}}^k = -\log 
\frac{\exp(\text{sim}(\mathbf{z}_P^k, \mathbf{z}_T^{y}) / \tau)}
{\sum_{c=1}^C \exp(\text{sim}(\mathbf{z}_P^k, \mathbf{z}_T^{c}) / \tau)},
\end{equation}
and aggregate the losses as:
\begin{equation}
\mathcal{L}_{\text{aug}} = \frac{1}{K} \sum_{k=1}^K \mathcal{L}_{\text{CE}}^k.
\label{eq:loss_aug}
\end{equation}

\begin{table*}[!t]
   \centering 
   \small
   \caption{\textbf{Performance across 5 cross-dataset benchmarks on OOD generalization on 3D Vision Language Models (VLMs).}  
   We report the overall accuracy (\%)  and standard deviation for the source and target domains, and report the average accuracy over the five target datasets in the last column.}
   \label{tab:ood}
   \setlength{\tabcolsep}{4pt}
   \small
   \resizebox{0.8\textwidth}{!}{
   \begin{tabular}{l c c c c c c c c}
      \toprule
      \multirow{2}{*}{Method} & \textbf{Source} & & \multicolumn{5}{c}{\textbf{Target}} & \multirow{2}{*}{\textbf{Avg.}} \\
             & ShapeNetV2 & & ModelNet40 & S-PB\_T50\_RS & S-OBJ\_BG & S-OBJ\_ONLY & Omni3D & \\
      \midrule
      P-CLIP~\citep{pointclip} & 67.41(0.09) & & 33.20(1.86) & 15.51(0.58) & 18.59(1.40) & 22.89(2.32) & 0.48(0.17) & 22.55 \\
      P-CLIP2~\citep{pointclip2} & 68.93(1.43) & & 54.73(1.48) & 39.53(4.22) & 34.30(1.28) & 25.63(1.16) & 8.63(2.52) & 32.56 \\
      \midrule
      ULIP~\citep{ulip} & 87.33(0.95) & & 56.17(1.15) & 26.83(2.15) & 39.43(2.17) & 43.53(1.32) & 6.37(0.90) & 34.47 \\
      PointPRC~\citep{pointprc} & 90.43(0.86) & & 58.00(0.57) & 28.43(0.68) & 40.33(0.71) & 46.33(1.54) & 8.20(0.50) & 36.26 \\
      \rowcolor{skyblue} \textbf{ReFine3D} & \textbf{91.56}(0.45) & & \textbf{60.66}(0.50) & \textbf{34.03}(0.53) & \textbf{45.18}(0.60) & \textbf{50.05}(0.98) & \textbf{12.47}(2.50) & \textbf{40.48} \\ %
      \midrule
      ULIP-2~\citep{ulip2} & 76.70(1.37) & & 65.27(0.66) & 40.07(0.34) & 53.80(1.78) & 48.53(1.72) & 17.27(0.54) & 44.99 \\ 
      PointPRC~\citep{pointprc} & 76.70(1.59) & & 72.10(0.93) & 46.77(2.43) & 59.03(3.02) & 56.27(0.97) & 21.80(0.49) & 51.19 \\ %
      \rowcolor{skyblue} \textbf{ReFine3D} & \textbf{80.08}(1.02) & & \textbf{75.01}(0.90) & \textbf{48.93}(2.03) & \textbf{62.01}(2.21) & \textbf{58.25}(0.59) & \textbf{22.89}(0.61) & \textbf{53.62} \\ %
      
      \bottomrule
   \end{tabular}
   }
\end{table*}

\begin{table*}
\small
\centering
   \caption{\textbf{Performance on corruption generalization. Here, ModelNet40 clean data is the source domain, and the target domain is ModelNet-C \citep{modelnetC}} with different types of corruption, with the corruption severity=2. }
   \label{tab:xset_corruption_generalization}
   \setlength{\tabcolsep}{5pt}
   \resizebox{0.8\textwidth}{!}{
   \begin{tabular}{l c c c c c c c c c}
      \toprule
      \multirow{2}{*}{Method} & \textbf{Clean Data} & \multicolumn{7}{c}{\textbf{Corruption Type}} & \multirow{2}{*}{\textbf{Avg.}} \\
             & ModelNet & Add Global & Add Local & Drop Global & Drop Local & Rotate & Scale & Jitter & \\
      \midrule
      P-CLIP~\citep{pointclip} & 80.97(1.02) & 80.97(1.02) & 80.97(1.02) & 64.95(1.08) & 68.31(1.93) & 65.75(1.19) & 72.04(1.33) & 52.09(1.28) & 69.30 \\
      P-CLIP2~\citep{pointclip2} & 83.49(0.51) & 83.49(0.51) & 83.49(0.51) & 68.85(3.22) & 66.67(1.96) & 70.13(1.33) & 75.68(0.15) & 61.21(2.16) & 72.79 \\ 
      \midrule
      ULIP~\citep{ulip} & 82.43(1.25) & 82.50(0.99) & 82.27(1.17) & 80.77(1.03) & 65.43(1.02) & 72.27(1.56) & 74.67(1.58) & 45.60(0.65) & 71.93 \\
      PointPRC~\citep{pointprc} & 83.87(0.34) & 83.83(0.40) & 83.93(0.19) & 81.83(0.52) & 67.37(1.72) & 79.10(0.36) & 76.37(0.09) & 41.67(4.79) & 73.44 \\ %
      \rowcolor{skyblue} \textbf{ReFine3D} & \textbf{84.93}(0.32) & \textbf{85.08}(0.33) & \textbf{84.53}(0.15) & \textbf{83.07}(0.41) & \textbf{70.73}(0.98) & \textbf{80.47}(0.35) & \textbf{77.73}(0.10) & \textbf{44.88}(2.26) & \textbf{75.21} \\
      \midrule
      ULIP-2~\citep{ulip2} & 85.07(0.21) & 81.97(0.79) & 82.03(0.96) & 79.93(0.92) & 60.03(1.21) & 80.30(0.93) & 75.77(0.74) & 44.27(2.13) & 72.04 \\ %
      PointPRC~\citep{pointprc} & 86.47(0.56) & 86.57(0.48) & 86.30(0.51) & 84.87(0.48) & 67.80(1.20) & 84.60(0.22) & 81.17(1.05) & 46.43(2.45) & 76.82 \\
      \rowcolor{skyblue} \textbf{ReFine3D} & \textbf{88.23}(0.50) & \textbf{88.18}(0.23) & \textbf{87.83}(0.45) & \textbf{87.01}(0.41) & \textbf{70.23}(0.98) & \textbf{86.97}(0.25) & \textbf{82.03}(0.95) & \textbf{48.08}(1.96) & \textbf{78.62} \\
      \bottomrule
   \end{tabular}
   }
\end{table*}
This encourages the model to produce consistent embeddings across augmentations, reducing memorization of specific input instances.

To reduce overfitting to class embeddings generated from class names with the text encoder, we propose a text synonymization strategy that generates diverse text prompts for each class using synonyms and a large language model (LLM). For a class $c$ with name $n_c$, we use WordNet to obtain a set of synonyms $\{s_c^1, s_c^2, \dots, s_c^M\}$. Additionally, we leverage an LLM to generate descriptive prompt sentences incorporating $n_c$ and its synonyms, resulting in a set of prompts $\{p_c^1, p_c^2, \dots, p_c^Q\}$. Each prompt is passed through the text encoder $f_T$ to obtain embeddings $\{\mathbf{z}_T^{c,q} = f_T(p_c^q)\}_{q=1}^Q$. Finally, the new class embedding is computed as the average:
\begin{equation}
\mathbf{z}_T^c = \frac{1}{Q} \sum_{q=1}^Q \mathbf{z}_T^{c,q}.
\end{equation}
The averaged class embeddings are used in the CE loss (Eq.~\ref{eq:1}), promoting robustness to variations in class descriptions.

Furthermore, to leverage the knowledge embedded in the pre-trained vision encoder, we introduce \textit{point-rendered vision supervision} to align point cloud embeddings with their rendered image embeddings. For each point cloud, we render a 2D image $\mathbf{I}$ using an off-the-shelf renderer (see details in Appendix). The image is passed through the frozen vision encoder $f_I$ to obtain an image embedding $\mathbf{z}_I = f_I(\mathbf{I})$. Here, we keep the vision encoder parameters fixed and learn only the point cloud encoder $f_P$ to align the augmented point cloud embedding $\mathbf{z}_P^k$ with the image embedding $\mathbf{z}_I$. We use a standard InfoNCE-style contrastive loss~\citep{oord2018representation}:
\begin{equation}
\mathcal{L}_{\text{contrast}}^k = -\log 
\frac{\exp(\text{sim}(\mathbf{z}_P^k, \mathbf{z}_I) / \tau)}
{\sum_{\mathbf{z}' \in \{\mathbf{z}_I\} \cup \mathcal{Z}_{\text{neg}}} 
\exp(\text{sim}(\mathbf{z}_P^k, \mathbf{z}') / \tau)},
\end{equation}
where $\mathcal{Z}_{\text{neg}}$ includes negative image embeddings from other point clouds in the batch, and $\text{sim}(\cdot, \cdot)$ denotes cosine similarity. The contrastive losses are averaged across augmentations similar to Eq.~\ref{eq:loss_aug}:
\begin{equation}
\mathcal{L}_{\text{contrast}} = \frac{1}{K} \sum_{k=1}^K \mathcal{L}_{\text{contrast}}^k.
\end{equation}

\begin{table*}[ht]
   \centering\scriptsize
   \caption{\textbf{Performance on cross-dataset generalization on \textit{PointDA}}\citep{pointdan}. M: ModelNet, S: ShapeNet, S*: ScanNet. We present the average across 6 evaluation setups in the last column.}
   \label{tab:xset_pointda_generalization}
   \begin{tabular}{l c c c c c c c}
      \toprule
      Method & M $\rightarrow$ S & M $\rightarrow$ S* & S $\rightarrow$ M & S $\rightarrow$ S* & S* $\rightarrow$ M 
      & S* $\rightarrow$ S & \textbf{Avg.} \\
      \midrule
      P-DAN~\citep{pointdan} & 64.2 & 33.0 & 47.6 & 33.9 & 49.1 & 64.1 & 48.7 \\
      MetaSets~\citep{metasets} & 86.0 & 52.3 & 67.3 & 42.1 & 69.8 & 69.5 & 64.5 \\
      PDG~\citep{pdg} & 85.6 & 57.9 & 73.1 & 50.0 & 70.3 & 66.3 & 67.2 \\
      I-OODG~\citep{zhang2024invariantoodg} & 83.7 & 56.4 & 71.7 & 57.6 & 69.5 & 73.5 & 67.8\\ %
      P-CLIP2~\citep{pointclip2} & 40.53 & 26.40 & 31.33 & 35.57 & 16.30 & 24.97 & 29.18 \\
      ULIP~\citep{ulip} & 74.33 & 38.23 & 35.17 & 36.17 & 24.70 & 60.67 & 44.88 \\
      \midrule
      ULIP-2~\citep{ulip2} & 84.80(2.69) & 48.10(2.13) & 83.20(4.17) & 42.00(4.18) & 60.43(4.83) & 70.50(6.22) & 64.84 \\ %
      PointPRC\citep{pointprc} & 89.00(1.18) & 51.37\textbf{(1.03)} & 89.87(2.38) & 49.57(2.50) & 85.57(3.80) & 83.07(4.21) & 74.74 \\ %
      \rowcolor{skyblue} \textbf{ReFine3D} & \textbf{90.22}\textbf{(1.05)} & \textbf{53.17}(1.11) & \textbf{92.07}\textbf{(2.02)} & \textbf{53.25}\textbf{(1.39)} & \textbf{87.31}\textbf{(2.23}) & \textbf{87.16}\textbf{(3.08)} & \textbf{77.19} \\
      \bottomrule
   \end{tabular}
\end{table*}

\begin{table*}[ht]
   \centering\scriptsize
   \caption{\textbf{Performance on cross-dataset generalization on \textit{Sim-to-Real}}\citep{metasets} in two evaluation settings, 
   MN\_11 $\rightarrow$ SONN\_11, SN\_9 $\rightarrow$ SONN\_9. }
   \resizebox{1.0\textwidth}{!}{
   \begin{tabular}{l c c c c c c c c}
      \toprule
      \multirow{2}{*}{Method} & \multicolumn{3}{c}{MN\_11 $\rightarrow$ SONN\_11} & & \multicolumn{3}{c}{SN\_9 $\rightarrow$ SONN\_9} & \multirow{2}{*}{\textbf{Avg.}} \\\cline{2-4}\cline{6-8}
             & OBJ & OBJ\_BG & PB\_T50\_RS & & OBJ & OBJ\_BG & PB\_T50\_RS\\
      \midrule
      MetaSets-P~\citep{metasets} & 60.3 & 52.4 & 47.4 & & 51.8 & 44.3 & 45.6 & 50.3 \\ %
      MetaSets-D~\citep{metasets} & 58.4 & 59.3 & 48.3 & & 49.8 & 47.4 & 42.7 & 51.0\\ %
      PDG-P~\citep{pdg} & 67.6 & 58.5 & 56.6 & & 57.3 & 51.3 & 51.3 & 57.1 \\ %
      PDG-D~\citep{pdg} & 65.3 & 65.4 & 55.2 & & 59.1 & 59.3 & 51.0 & 59.2 \\ %
      P-CLIP2~\citep{pointclip2} & 18.67 & 15.57 & 15.63 &  & 53.00 & 47.83 & 35.83 & 31.09 \\
      ULIP~\citep{ulip} & 21.60 & 18.03 & 13.63 & & 54.83 & 54.17 & 40.87 & 33.86 \\
       \midrule
      ULIP-2~\citep{ulip2} & 62.73(0.95) & 68.23(0.86) & 52.83(1.10) & & 66.90(2.77) & 70.50(2.48) & 54.03(2.75) & 62.54 \\ %
      PointPRC \citep{pointprc} & 68.43(1.07) & 69.47(0.95) & 55.30\textbf{(2.00)} & & 65.83(1.35) & 72.53\textbf{(0.47)} & 58.57(1.17) & 65.02 \\ %
    \rowcolor{skyblue} \textbf{ReFine3D} & \textbf{70.11}\textbf{(0.91)} & \textbf{72.53}\textbf{(0.78)} & \textbf{56.87}(2.15) && \textbf{66.92}\textbf{(1.22)} & \textbf{73.03}(0.62) & \textbf{60.00}\textbf{(0.52)} & \textbf{66.58} \\
      \bottomrule
   \end{tabular}}
   \label{tab:xset_sim2real_generalization}
\end{table*}

 \vspace{-10pt}

This pre-trained point cloud encoder is then used for fine-tuning on downstream tasks, where we align point cloud embeddings $\mathbf{z}_P$ with text embeddings $\mathbf{z}_T^{y}$ using cosine similarity and the CE loss (Eq.~\ref{eq:1}). 
Aligning point cloud features with both rendered image and text embeddings during fine-tuning allows the model to jointly leverage spatial cues from vision and semantic structure from language, leading to more discriminative and robust point cloud representations. Finally, the total loss function optimized jointly over all augmentations is:
\begin{equation}
\mathcal{L} = \frac{1}{K} \sum_{k=1}^K 
\left( \mathcal{L}_{\text{CE}}^k + \alpha \cdot \mathcal{L}_{\text{contrast}}^k \right),
\end{equation}
where $\alpha$ is a hyperparameter to balance the contributions of the contrastive and classification losses.

To enhance generalization at inference, we introduce a test-time augmentation mechanism with confidence-based aggregation inspired by leveraging additional computation at inference~\citep{wei2022chain, turpin2023language}. Unlike training-free test-time adaptation methods~\citep{sun2025point} that operate in zero-shot settings, our approach leverages the task-specific knowledge acquired during few-shot fine-tuning. For a test point cloud $\mathbf{x}_{\text{test}}$, we generate $K_{\text{test}}$ augmentations $\{\mathbf{x}_{\text{test}}^k\}_{k=1}^{K_{\text{test}}}$ and compute their embeddings $\{\mathbf{z}_P^k = f_P(\mathbf{x}_{\text{test}}^k)\}_{k=1}^{K_{\text{test}}}$. 
On the text branch, we use the synonymization strategy to generate multiple class embeddings $\{\mathbf{z}_T^{c,q}\}_{q=1}^Q$ per class and aggregate results via averaging. Then we compute similarities of 
these text embeddings with the augmented point cloud embeddings. For each class, we select the top-$H$ embeddings with the highest confidence (based on maximum softmax probability) and aggregate their predictions via majority voting. This approach leverages additional computation to improve robustness and accuracy by reducing erroneous predictions during inference.

\subsection{Implementation Details}
\label{subsec:impl_details}
We adopt ULIP-2~\citep{ulip2} as the pre-trained backbone, since it is the most widely adopted pre-trained foundation model in the existing fine-tuning literature. We perform all experiments using three different random seeds and report the mean and standard deviation to ensure statistical reliability. 
For model training, we use stochastic gradient descent (SGD) as the optimizer with an initial learning rate of 0.0025. We present a detailed hyperparameter configuration in Table \ref{tab:hyperparameters}.
\vspace{-10pt}

\subsection{Details on Image Rendered}
\vspace{-10pt}
Rendering images from point clouds is a one-time offline operation independent of the actual training process. Although we generated the rendered images for our method, many existing datasets already provide rendered images alongside 3D data (e.g., widely used 3D-vision benchmarks). In our implementation, rendering has a throughput of approximately 100 samples per second using Blender and BlenderProc \citep{denninger2019blenderproc}.
ModelNet40 and ShapeNetCoreV2 sampled to 10,000 points via uniform sampling to align with standard practices \citep{pointnet}, while ScanObjectNN and Objaverse retain native point densities (2,000–50,000 points) to preserve real-world noise and complex geometries, incorporating surface normals and RGB attributes when available. 
The rendering configuration is optimized for consistency: a directional light (intensity 1.0, 45° elevation, 30° azimuth) simulates natural sunlight, paired with an ambient light (intensity 0.25) to soften shadows. The camera employs a perspective projection with a 35mm focal length, 32mm sensor width, and 512x512 pixel resolution, with extrinsics placing the camera 1.8 meters from the object’s centroid across 12 viewpoints (azimuthal angles 0°–360° in 30° increments, elevation angles 0°, 15°, 30°). Materials use a Lambertian diffuse shader (reflection coefficient 0.8) based on point cloud normals or RGB data, ensuring realistic shading for synthetic and real-world objects without specular highlights. A solid gray background (RGB 0.5, 0.5, 0.5) maintains focus on objects, aligning with common practices in ShapeNetCoreV2 and ModelNet40.

\begin{wraptable}{r}{0.45\textwidth}
\vspace{-20pt}
    \centering\scriptsize
        \caption{\textbf{Performance across 5 benchmarks on few-shot generalization.} We report the overall accuracy and the average on the different shots ranging from 1,2,4,8, and 16. }
    \label{tab:fewshot}
    \setlength{\tabcolsep}{2pt}
    \begin{tabular}{c|l|cccccc}
\toprule[1.1pt]
\rot{\textbf{Setting}} &  \rot{\textbf{Methods}} & \textbf{\rot{Average}}&  \textbf{\rot{ModelNet40}} & \textbf{\rot{S-PB\_T50\_RS}} &  \textbf{\rot{S-OBJ\_BG}} &  \textbf{\rot{S-OBJ\_ONLY}} & \textbf{\rot{Omni3D}} \\
\cmidrule{1-8}

\multirow{2}{*}{1-shot} & PointPRC &  {63.49} & {65.53} & {52.10} & {64.50} & {64.80} & {70.50} \\
 & \ours ReFine3D & {66.60} &  {67.87} & {55.78} & {67.48} & {69.57} & {72.29} \\ \midrule

\multirow{2}{*}{2-shot}& PointPRC & {66.88} & {68.41} & {53.50} & {72.00} & {69.50} & {71.00} \\
 & \ours ReFine3D & {68.18}&  {71.13} & {55.32} & {72.18} & {69.84} & {72.43}  \\ \midrule

\multirow{2}{*}{4-shot}& PointPRC & {72.21} &  {70.05} & {68.50} & {73.50} & {75.00} & {74.00}   \\
 & \ours ReFine3D &  {75.14}&  {73.64} & {70.96} & {76.44} & {76.16} & {78.51}   \\ \midrule

\multirow{2}{*}{8-shot} & PointPRC &  {72.77} & {72.34} & {63.50} & {76.00} & {78.00} & {74.00} \\
 & \ours ReFine3D & {75.82} &  {75.02} & {68.15} & {77.91} & {77.93}& {80.08} \\ \midrule

\multirow{2}{*}{16-shot} & PointPRC &  {78.92} & {77.17} & {73.97} & {84.28} & {82.33} & {76.85} \\
 & \ours ReFine3D & {80.44} &  {78.34} & {75.90} & {86.28} & {83.51} & {78.19} \\ 
\bottomrule[1.1pt]
\end{tabular}
\end{wraptable}

\vspace{10pt}

\section{Results and Experiments}
\label{sec:result}

\subsection{Evaluation Protocol}
\label{subsec:datasets}

To ensure fair comparison with existing approaches in 3D domain generalization (3D-DG), we adopt the evaluation protocol established by prior work \citep{pointprc}. This protocol includes three key benchmarks: base-to-new generalization, cross-dataset generalization, and few-shot learning. The base-to-new benchmark is conducted on five widely used datasets—ModelNet40~\citep{modelnet}, ShapeNetCoreV2~\citep{shapenet}, and three variants of ScanObjectNN~\citep{scanobject} (S-PB\_T50\_RS, S-OBJ\_BG, and S-OBJ\_ONLY). Each dataset is evenly divided into base and novel classes. Fine-tuning is performed on the base classes, while model generalization is evaluated on novel classes. For cross-dataset generalization, we consider four settings: (i) out-of-distribution (OOD) generalization, where models trained on one domain are evaluated on different domains; (ii) data corruption, where training is conducted on clean ModelNet40 and testing is done on its corrupted counterpart, ModelNet-C~\citep{modelnetC}; (iii) domain adaptation following PointDA~\citep{pointdan}; and (iv) sim-to-real transfer using the protocol introduced in~\citep{metasets}. Additionally, to assess performance under limited supervision, we conduct few-shot learning experiments with 1, 2, 4, 8, and 16 labelled samples per class from the base set, and evaluate on the full test set. If not specified, we perform the experiments on 16-shot setting. 

\subsection{Base-to-new Class Generalization}
We evaluate our proposed method on the base-to-new class generalization setting across five 3D benchmarks, as summarized in Table~\ref{tab:base2new}. The results demonstrate that ReFine3D consistently outperforms existing methods across all metrics, including base accuracy, new class accuracy, and their harmonic mean (HM). Specifically, ReFine3D improves the generalization performance on both novel and base classes, achieving the highest new class accuracy of 77.46\% and base class accuracy of 84.71\% on average.

Notably, we observe the largest improvement in HM on datasets with higher variability and more challenging semantic gaps between base and novel classes, such as S-PB\_T50\_RS with HM gains of 1.93\% over PointPRC~\citep{pointprc}. ReFine3D also shows strong and consistent performance on ModelNet40 and S-OBJ\_BG and ShapeNetCoreV2, confirming the method’s robustness across diverse scenarios. On average across the five datasets, ReFine3D achieves an HM of 80.40\%, representing a relative gain of 0.92\% over the previous state-of-the-art PointPRC.


\begin{table}[t]
    \centering
    \small
    \caption{\textbf{Pairwise comparison between ReFine3D and PointPRC.} We report the percentage of samples where both methods are correct, only ReFine3D is correct, only PointPRC is correct, or both methods are wrong. Win Rate is computed as $\frac{\text{ReFine3D Only}}{\text{ReFine3D Only}+\text{PointPRC Only}}$.}
    \label{tab:pairwise_comparison}
    \begin{tabular}{l|c c c c c}
        \toprule
        \textbf{Dataset} & \textbf{Both Correct} & \textbf{ReFine3D Only} & \textbf{PointPRC Only} & \textbf{Both Wrong} & \textbf{Win Rate} \\
        \midrule
        ModelNet40  & 63.2 & 15.1 & 11.1 & 10.6 & \textbf{57.6} \\
        S-PB-T50-RS & 51.8 & 24.0 & 22.2 & 2.0  & \textbf{51.9} \\
        S-OBJ-BG    & 52.1 & 24.8 & 23.1 & 0.0  & \textbf{51.8} \\
        S-OBJ-ONLY  & 65.7 & 17.6 & 16.7 & 0.0  & \textbf{51.3} \\
        ShapeNetV2  & 70.5 & 15.9 & 13.6 & 0.0  & \textbf{53.9} \\
        \midrule
        \rowcolor{skyblue}
        \textbf{Average} & \textbf{60.7} & \textbf{19.5} & \textbf{17.3} & \textbf{2.5} & \textbf{53.3} \\
        \bottomrule
    \end{tabular}
\end{table}

\subsection{Cross-Dataset Generalization}

In this section, we analyze our proposed method's performance for OOD generalization and data corruption settings. We present a detailed analysis of the other two settings, domain adaptation \citep{pointdan} and sim-to-real \citep{metasets}.

\begin{wraptable}{r}{0.25\textwidth}
    \vspace{-15pt}
    \centering
    \small
    \caption{\textbf{McNemar's test on paired disagreements.}}
    \label{tab:mcnemar_test}
    \begin{tabular}{l|c}
        \toprule
        \textbf{Dataset} & \textbf{p-value} \\
        \midrule
        ModelNet40  & \textbf{0.031} \\
        S-PB-T50-RS & \textbf{0.042} \\
        S-OBJ-BG    & \textbf{0.038} \\
        S-OBJ-ONLY  & \textbf{0.047} \\
        ShapeNetV2  & \textbf{0.029} \\
        \bottomrule
    \end{tabular}
    \vspace{-10pt}
\end{wraptable}

\textbf{OOD generalization.}
Out-of-distribution (OOD) generalization assesses a model’s capacity to transfer knowledge learned from a known domain to unseen target domains. In this benchmark, we use ShapeNetV2 as the source domain and evaluate the model's generalization on five diverse target datasets. 
As shown in Table~\ref{tab:ood}, ReFine3D significantly outperforms prior 3D vision-language models in both source and target domains. Our method improves existing SOTA by +3.38\% on the source domain with an average of +2.43\% across 5 target domains.

\textbf{Data corruption.} 
In the real world, point clouds often suffer from data corruption due to sensor noise, incomplete scans, and geometric irregularities. To evaluate the robustness of our framework under such realistic perturbations, we test on ModelNet-C~\citep{modelnetC}, which applies seven common corruptions at severity level 2. As shown in Table~\ref{tab:xset_corruption_generalization}, ReFine3D achieves the highest accuracy on both clean and corrupted data settings. On the clean ModelNet40 dataset, ReFine3D improves SOTA by +1.76\%. When averaged across 7 corruptions, ReFine3D yields a substantial +1.80\% gain, consistently leading in all corruption types. 

\begin{wraptable}{r}{0.5\textwidth}
    \centering\small
    \vspace{-10pt}
    \caption{\textbf{Ablation study on the different fine-tuning strategies of our proposed framework.} We report the overall accuracy and harmonic mean (HM) on the Base-to-new generalization setting on S-PB\_T50\_RS.} 
    \label{tab:ablate_base2new}
    \begin{tabular}{ c | c c c}
       \toprule
       \textbf{Method}& \textbf{Base} & \textbf{New} & \textbf{HM} \\
       \midrule
       \rowcolor{skyblue}ReFine3D (ours) & 76.00 & 75.80 & 75.90 \\ %
       w/o vision guidance & 73.98 & 74.50 & 74.24 \\ %
       w/o test-time augmentation & 74.06 & 74.93 & 74.50 \\\ %
       w/o layer-specific fine-tuning & 74.12 & 74.58 & 74.35 \\ %
       w/o training augmentation & 74.25 & 74.62 & 74.43 \\ %
       w/o text synonymization & 74.18 & 74.06 & 74.12 \\ %
       \bottomrule
    \end{tabular}
     \vspace{-5pt}
\end{wraptable}
\textbf{Domain adaptation.}
PointDA, a 3D domain adaptation benchmark pioneered by PointDAN \citep{pointdan}, features six evaluation scenarios outlined in Table \ref{tab:xset_pointda_generalization}. Unlike earlier techniques such as MetaSets \citep{metasets}, PDG \citep{pdg}, and I-OODG \citep{zhang2024invariantoodg}, which utilize the complete training dataset across each scenario, our ReFine3D method relies on a few-shot learning approach with 16 samples. The outcomes underscore ReFine3D's strong performance in domain adaptation, delivering an average accuracy of 77.19\% across all setups. Remarkably, it outpaces the existing SOTA method, PointPRC, by 2.45\%, with notable gains including a 4.09\% boost when the source domain is ScanNet and target domain is ShapeNet, and a 3.68\% rise on ShapeNet to ScanNet generalization, highlighting its effectiveness in improving cross-dataset generalization with limited data.

\textbf{Sim-to-Real generalization.}
The Sim-to-Real evaluation assesses cross-domain generalization by transitioning from simulated to real-world data, a concept initially explored by MetaSets \citep{metasets} and 
\begin{wraptable}{r}{0.35\textwidth}
    \centering\small
    \caption{\textbf{Comprehensive ablation of all component combinations in ReFine3D.} PA = point cloud augmentation, TA = test-time augmentation, TS = text synonymization, VG = vision guidance. We report harmonic mean (HM) on Base-to-New generalization of S-PB-T50-RS.}
    \setlength{\tabcolsep}{2pt}
    \begin{tabular}{cccc|c|l}
    \toprule
    \textbf{PA} & \textbf{TA} & \textbf{TS} & \textbf{VG} & \textbf{HM} & \textbf{Label} \\
    \midrule
    \xmark & \xmark & \xmark & \xmark & 71.67 & All Off \\
    \cmark & \xmark & \xmark & \xmark & 72.63 & Only PA \\
    \xmark & \cmark & \xmark & \xmark & 72.75 & Only TA \\
    \xmark & \xmark & \cmark & \xmark & 72.77 & Only TS \\
    \xmark & \xmark & \xmark & \cmark & 72.97 & Only VG \\
    \cmark & \cmark & \xmark & \xmark & 73.50 & PA+TA \\
    \cmark & \xmark & \cmark & \xmark & 73.59 & PA+TS \\
    \cmark & \xmark & \xmark & \cmark & 73.42 & PA+VG \\
    \xmark & \cmark & \cmark & \xmark & 73.62 & TA+TS \\
    \xmark & \cmark & \xmark & \cmark & 73.52 & TA+VG \\
    \xmark & \xmark & \cmark & \cmark & 73.44 & TS+VG \\
    \cmark & \cmark & \cmark & \xmark & 74.43 & w/o PA \\
    \cmark & \cmark & \xmark & \cmark & 74.35 & w/o TS \\
    \cmark & \xmark & \cmark & \cmark & 74.50 & w/o TA \\
    \xmark & \cmark & \cmark & \cmark & 74.24 & w/o VG \\
    \cmark & \cmark & \cmark & \cmark & 75.90 & \textbf{ReFine3D} \\
    \bottomrule
    \vspace{-60pt}
    \end{tabular}
    \label{tab:comprehensive_ablation}
\end{wraptable}
later expanded by PDG \citep{pdg}. In this evaluation, ModelNet \citep{modelnet} and ShapeNet \citep{shapenet} serve as synthetic point cloud sources, while ScanObjectNN \citep{scanobject} is derived from real-scanned data. 
Unlike MetaSets and PDG, which rely on the entire training set in the source domain for supervised learning, our approach utilizes only 16-shot prompt tuning. As shown in Table \ref{tab:xset_sim2real_generalization}, our framework consistently boosts generalization across various 3D models, with an improvement of 1.56\% over the SOTA method averaged over six datasets.

\subsection{Few-shot Generalization}
To assess the effectiveness of ReFine3D in label-constrained scenarios, we evaluate its performance across five benchmarks under varying few-shot settings.
As shown in Table~\ref{tab:fewshot}, ReFine3D consistently outperforms existing SOTA across all shots and benchmarks. The performance gains are especially notable in extremely low-shot scenarios. For instance, ReFine3D achieves an average improvement of 3.11\% in the 1-shot setting. 
Notably, performance continues to improve steadily with more labelled samples, reaching 80.44\% at 16 shots. This trend underscores the robustness and adaptability of our approach as label availability increases.

\subsection{Sample-level Analysis }
We report a paired, sample-level disagreement analysis between ReFine3D and PointPRC (Table \ref{tab:pairwise_comparison}). Rather than relying solely on aggregate accuracy, we have analyzed predictions on a per-sample basis and partitioned the test set into four mutually exclusive categories: (1) both correct, (2) ReFine3D only correct, (3) PointPRC only correct, and (4) both wrong. This paired evaluation isolates the exact samples on which the methods disagree and directly measures relative superiority.
Across all five benchmarks, ReFine3D wins more disagreements than it loses. On average, ReFine3D correctly classifies 19.5\% of samples that PointPRC misclassified, compared to 17.3\% vice versa, yielding an average win rate of 53.3\%. Importantly, ReFine3D achieves a win rate greater than 50\% on every dataset, demonstrating consistent per-sample advantage rather than isolated gains.
\begin{wraptable}{r}{0.55\textwidth}
    \centering 
    \small
    \vspace{-10pt}
        \caption{\textbf{Ablation study on tuning different layers of point cloud encoder.} We report the overall accuracy and harmonic mean (HM) on the Base-to-new generalization setting on S-PB\_T50\_RS.} 
    \label{tab:ablate_layers}
    \begin{tabular}{ c | c c c}
       \toprule
       \textbf{Number of layers} & \textbf{Base} & \textbf{New} & \textbf{HM} \\
       \midrule
       \rowcolor{skyblue}Last layer attention block only & 76.00 & 75.80 & 75.90 \\ %
       Last layers (last 4 attention blocks) & 75.82 & 75.14 & 75.48 \\ %
       Middle layers (3-7 attention block) & 74.96 & 74.34 & 74.65 \\ %
       Early layers (First 4 attention blocks)  & 73.15 & 74.02 & 73.58 \\ %
       Full finetuning  & 73.06 & 70.57 & 71.79 \\ %
       \bottomrule
    \end{tabular}
\end{wraptable}

To formally assess whether these disagreement asymmetries are statistically significant, we performed McNemar’s test on the paired disagreement counts (ReFine3D Only vs. PointPRC Only). The reported p-values evaluate whether the difference in prediction outcomes between the two methods is statistically significant. Lower p-values indicate stronger evidence that the observed performance difference is not due to chance. As shown in the Table \ref{tab:mcnemar_test}, all datasets yield p-values below 0.05, indicating that the observed asymmetries are unlikely to arise from random variation.

\subsection{Ablation and Sensitivity Study}

\textbf{Different fine-tuning strategies of ReFine3D.}
We present an ablation study on the different fine-tuning strategies of our proposed method, ReFine3D, in Table~\ref{tab:ablate_base2new}. ReFine3D achieves the highest performance with a harmonic mean (HM) of 75.90\% when all proposed tuning strategies are applied. Removing each component individually results in a consistent drop in performance, highlighting their collective significance. Notably, disabling vision guidance leads to the most significant decline, with HM dropping by 1.66\%. Text synonymization also contributes substantially, with its removal reducing HM by 1.78\%. Other tuning strategies---test-time 
scaling, layer-specific fine-tuning, and training augmentation---each contributes approximately 1.4–1.5\% to the overall performance when removed.


Table \ref{tab:comprehensive_ablation} demonstrate that the four fine-tuning strategies interact synergistically. Each component individually provides consistent improvement, yet when combined, their contributions compound substantially. Specifically, ReFine3D achieves 4.23\% overall improvement over the baseline, which exceeds both the sum of individual effects and any pairwise combination, indicating strong synergy among all components.


\begin{wraptable}{r}{0.6\textwidth}
\centering
\caption{\textbf{Ablation of test-time inference strategies on S-PB T50 RS.} Here, TTA represents test-time augmentation.}
\label{tab:abl_tta}
\begin{tabular}{l|ccc}
\toprule
\textbf{Test-Time Strategy} & \textbf{Base} & \textbf{New} & \textbf{HM} \\
\midrule
Single view (no scaling) & 74.12 & 74.58 & 74.35 \\
Simple TTA (avg. 5 views) & 74.80 & 75.02 & 74.91 \\
TTA + text diversity & 75.33 & 75.48 & 75.40 \\
\rowcolor{skyblue}Full Test-time augmentation (ours) & \textbf{76.00} & \textbf{75.80} & \textbf{75.90} \\
\bottomrule
\end{tabular}
\end{wraptable}

\textbf{Number of fine-tuned layers}. 
Table \ref{tab:ablate_layers} presents an ablation study on the layer-selective fine-tuning strategy. In this study, we keep the vision and text encoder frozen while selectively tuning different layers of the 3D encoder (e.g. PointBERT). We tune the layers at different levels, such as early, mid and last layers. The table shows that tuning the last layers results in the highest performance gain, indicating the significance of high-level semantic adaptation for generalization. On the contrary, the early and mid layers encode general low-level geometric patterns, thus contributing less to transferring knowledge in downstream tasks.

\begin{wraptable}{r}{0.4\textwidth}
    \centering\small
    \caption{\textbf{Ablation study on text prompts generated from different sources.} We report the overall accuracy and harmonic mean (HM) on the Base-to-new generalization setting on S-PB\_T50\_RS. Here, GPT-3.5 represents GPT-3.5-turbo.} 
    \label{tab:ablation_prompts}
    \begin{tabular}{ c | c c c}
       \toprule
       \textbf{Text propmts} & \textbf{Base} & \textbf{New} & \textbf{HM} \\
       \midrule
       Manual  & 73.98 & 74.50 & 74.24 \\ %
       Qwen-2.5  & \textbf{76.00} & \textbf{75.80} & \textbf{75.90} \\ 
      LLaMa-3  & 74.06 & 74.93 & 74.50 \\ %
       Mistral & 74.12 & 74.58 & 74.35 \\ %
       \bottomrule
    \end{tabular}
     \vspace{-10pt}
\end{wraptable}
\textbf{Test-time inference strategies}. 
We analyze the effectiveness of our test-time augmentation mechanism by comparing different inference strategies. Table \ref{tab:abl_tta} shows that simple test-time augmentation (averaging predictions over 5 views) provides modest gains (+0.56\% HM), while adding text diversity improves further (+1.05\% HM). Our full test-time augmentation, which combines augmentation, text diversity, and top-$H$ confidence-based selection, achieves the best performance (+1.55\% HM over single-view inference). Note that our test-time augmentation addresses a different problem than training-free test-time adaptation methods like Point-Cache~\citep{sun2025point}, which operate without any labelled downstream data. Point-Cache achieves robustness through dynamic feature caching in pure zero-shot settings, while our approach leverages task-specific knowledge from few-shot fine-tuning.

\begin{wrapfigure}{r!}{0.38\textwidth}
\vspace{-5pt}
\begin{center}
    \centering
    \includegraphics[width=1.0\linewidth]{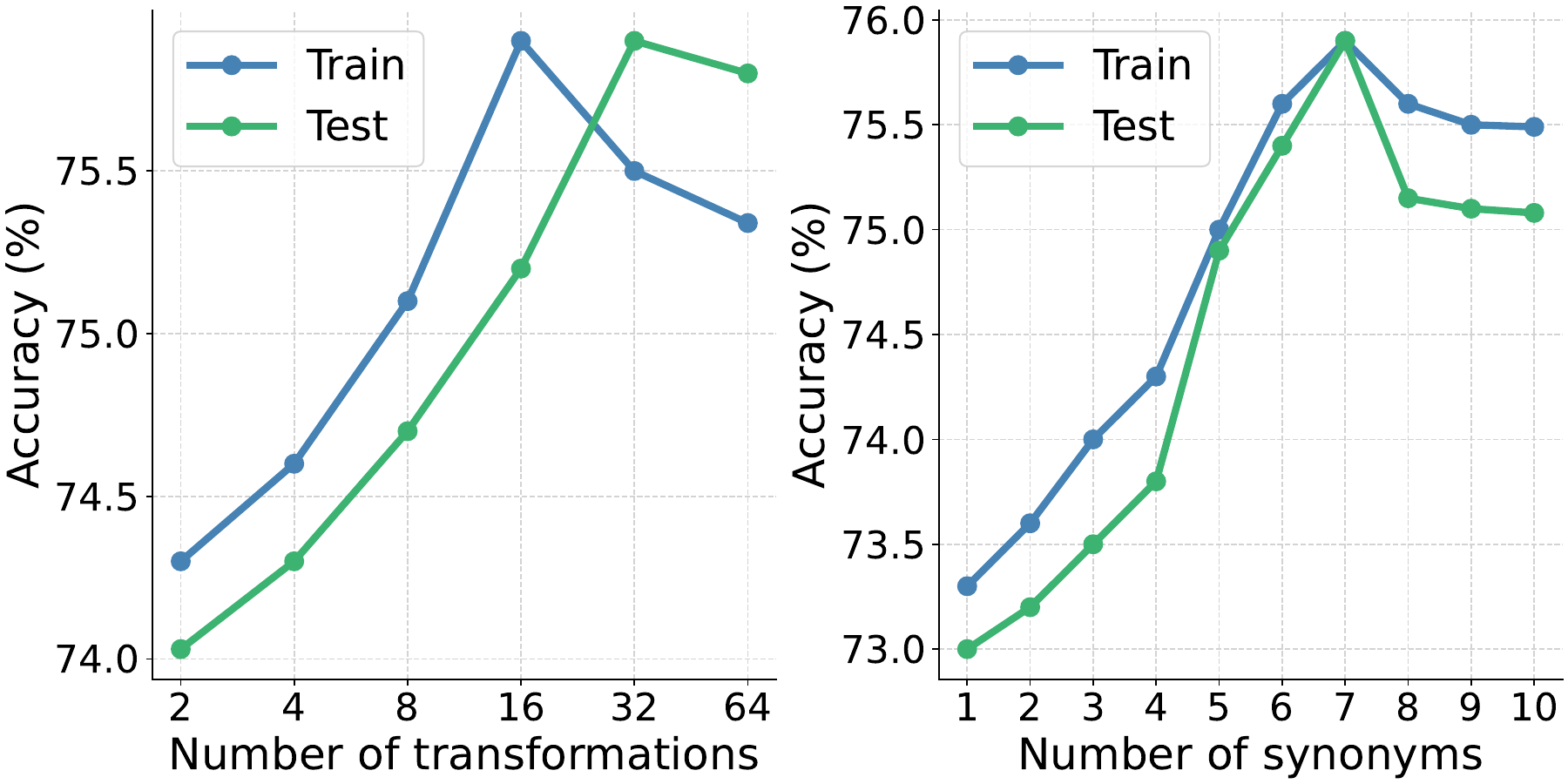}
    \caption{The impact of transformation strength and number of text synonyms during train and test time.}
    \label{fig:sensitivity}
\end{center}
 \vspace{-10pt}
\end{wrapfigure}
\textbf{Text prompts from various sources.}
\begin{table}[t]
    \centering\small
    \caption{\textbf{Qualitative examples of WordNet synonyms and LLM-generated prompts.} For each class, we show WordNet synonyms and corresponding LLM-generated descriptive prompts.}
    \setlength{\tabcolsep}{2pt} 
    \begin{tabular}{p{1.2cm}|p{4.8cm}|p{5.5cm}}
    \toprule
    \textbf{Class} & \textbf{WordNet Synonyms} & \textbf{LLM-Generated Prompts} \\
    \midrule
    \multirow{3}{*}{Airplane} & \multirow{3}{*}{aircraft, plane, jet} & A 3D point cloud of an aircraft with wings and a fuselage \\
    & & A 3D point cloud of a plane designed for air travel \\
    & & A 3D point cloud of a jet with engines mounted on the body \\
    \midrule
    \multirow{3}{*}{Chair} & \multirow{3}{*}{seat, furniture, seating} & A 3D point cloud of a seat with a backrest and legs \\
    & & A 3D point cloud of furniture designed for sitting \\
    & & A 3D point cloud of seating with armrests and a base \\
    \midrule
    \multirow{3}{*}{Lamp} & \multirow{3}{*}{light, lighting fixture, luminaire} & A 3D point cloud of a light source with a shade and stand \\
    & & A 3D point cloud of a lighting fixture for illumination \\
    & & A 3D point cloud of a luminaire with electrical components \\
    \bottomrule
    \end{tabular}
    \label{tab:qualitative_prompts}
\end{table}
Table \ref{tab:ablation_prompts} presents an ablation study comparing text prompts generated from different sources---Manual, Qwen-2.5, LLaMa-3, and Mistral---on a Base-to-New generalization task. Among these, Qwen-2.5 prompts deliver the highest results across all metrics (Base: 76.00, New: 75.80, HM: 75.90), clearly outperforming manually crafted prompts. This suggests that high-capacity language models can generate richer or more relevant textual cues. When compared to other LLMs, Qwen-2.5 also surpasses LLaMa-3 and Mistral, which still improve over manual prompts. 
Overall, these findings demonstrate that leveraging advanced LLM-generated prompts can significantly enhance the generalization capability of vision-language models on both seen and unseen categories.

To provide transparency in our text diversification pipeline, we include qualitative examples of WordNet synonyms and LLM-generated prompts used in ReFine3D. Table~\ref{tab:qualitative_prompts} shows examples from three representative classes with their WordNet synonyms and corresponding LLM-generated descriptive prompts using Qwen-2.5-7B-Instruct. These examples demonstrate the semantic diversity captured through our synonymization strategy, which helps the model learn more robust and generalizable representations by avoiding overfitting to specific class name phrasings.

\begin{wraptable}{r}{0.5\textwidth}
 \vspace{-20pt}
    \centering\small
    \caption{\textbf{Sensitivity study on different numbers of rendered images.} We report the overall accuracy and harmonic mean (HM) on the Base-to-new generalization setting on S-PB\_T50\_RS.} 
    \label{tab:ablation_views}
    \begin{tabular}{ c | c c c}
       \toprule
       \textbf{Number of views} & \textbf{Base} & \textbf{New} & \textbf{HM} \\
       \midrule
        1 & 76.00 & \textbf{75.80} & 75.90 \\ %
        2 & 77.46 & 75.68 & 76.56 \\ %
        3 & 79.23 & 74.10 & \textbf{76.57} \\ %
        4 & 80.10 & 72.00 & 75.78 \\ %
        5 & \textbf{80.60} & 70.30 & 74.72 \\
       \bottomrule
    \end{tabular}
\end{wraptable}

\textbf{Transformation strength.} Figure~\ref{fig:sensitivity} \textit{(left)} presents an analysis of the ReFine3D's performance during training and test time under various transformation settings. The left plot displays model performance under different numbers of geometric augmentations applied to point clouds. During training, accuracy peaks when 16 geometric augmentations are applied to the point cloud. During test time, the highest performance is achieved when 32 augmentations are applied.

Figure~\ref{fig:sensitivity} \textit{(right)} presents the model's performance with varying numbers of synonyms used in the text prompts. At both train and test time, 7 synonyms per word applied to the text prompts give the highest performance, indicating that the severity of transformations is critical to the model's performance. 


\textbf{Vision Supervision Trade-off Analysis.} Table \ref{tab:ablation_views} presents an ablation study evaluating the number of rendered image views affecting model performance under the Base-to-New generalization setting. Initially, increasing the number of views from 1 to 3 improves base class accuracy (from 76.00 to 79.23) by providing richer visual information. However, novel class accuracy declines, reflecting reduced generalization. As the number of views increases further (to 4 and 5), base accuracy continues to rise due to stronger memorization of base examples (reaching 80.60 at 5 views), but novel accuracy drops more noticeably (to 70.30 at 5 views), suggesting view-specific overfitting.

\begin{wraptable}{r}{0.35\textwidth}
\centering\small
\caption{\textbf{Sensitivity study on the loss balancing hyperparameter 
$\alpha$.} We report accuracy and harmonic mean (HM) under the Base-to-New generalization setting.}
\label{tab:ablation_loss}
\begin{tabular}{c|c c c}
\toprule
\textbf{$\alpha$} & \textbf{Base}  & \textbf{New}  & \textbf{HM} \\
\midrule
0.5 & 75.10 & 74.20 & 74.65 \\
1.0 & \textbf{76.00} & \textbf{75.80} & \textbf{75.90} \\
1.5 & 75.80 & 74.00 & 74.89 \\
\bottomrule
\end{tabular}
 \vspace{-15pt}
\end{wraptable}
This counterintuitive trade-off arises from view-specific overfitting. With multiple rendered views, the model learns to align point cloud features with view-dependent visual patterns (specific lighting, camera angles, surface normals) that are abundant in base class training data but absent from novel classes. Unlike geometric augmentations, which preserve object identity, multiple rendered views create consistent but view-specific visual anchors that the model memorizes rather than learning view-invariant 3D representations. Through our contrastive vision supervision loss, using a single view forces the encoder to extract the most salient and transferable 3D-to-2D mappings, whereas multiple views allow the encoder to overfit by learning all view-specific patterns simultaneously.

Our choice of 1 rendered view is an informed design decision that achieves three objectives: (1) balanced generalization with minimal base-novel gap (0.20\% vs. 10.30\% at 5 views), (2) sufficient visual supervision---removing vision guidance drops HM by 1.66\%, the largest single-component penalty, and (3) computational efficiency. This finding validates that vision supervision quality matters more than quantity, and strategic constraint in multimodal alignment is essential for maintaining generalization under domain shift.

\begin{wraptable}{r}{0.3\textwidth}
\vspace{-15pt}
\centering
\small
\caption {\textbf{Comparison between different LoRA configurations. }Results are reported in terms of harmonic mean (HM).}
\label{tab:ablation_hm}
\setlength{\tabcolsep}{4pt}
\begin{tabular}{l c}
\toprule
\textbf{Method} & \textbf{HM} \\
\midrule
PointPRC                & 73.97 \\
LoRA (Rank 4)           & 71.26 \\
LoRA (Rank 8)           & 74.00 \\
LoRA (Rank 12)          & 73.03 \\
LoRA (Attention Only)   & 72.82 \\
LoRA (MLP Only)         & 74.00 \\
LoRA (Attn + MLP)       & 71.08 \\
\rowcolor{skyblue}
\textbf{ReFine3D}       & \textbf{75.90} \\
\bottomrule
\end{tabular}
\vspace{-10pt}
\end{wraptable}

\textbf{Loss balancing hyperparameter, ($\alpha$).}
Table \ref{tab:ablation_loss} presents an ablation study on the loss balancing hyperparameter, $\alpha$, which controls the relative contribution of the contrastive loss in the final objective function. The results show that equal importance to both loss components optimally supports balanced learning and transferability and yields the best overall performance. Lower or higher values result in decreased HM scores, suggesting that under- or over-weighting the contrastive loss degrades generalization.

\textbf{LoRA baselines.}
To validate our layer-selective strategy against standard PEFT alternatives, 
we compare ReFine3D with LoRA~\citep{hu2022lora} under varying rank 
configurations and insertion locations (Table~\ref{tab:ablation_hm}). 
LoRA performance varies considerably: while Rank~8 and MLP-only match 
PointPRC (74.00), other configurations underperform it, with Attn+MLP 
dropping to 71.08---reflecting the risk of disrupting pre-trained geometric 
representations in early layers critical for cross-domain transfer.

\subsection{Computational Complexity}
\label{sec:run_time}
\begin{wraptable}{r}{0.38\textwidth}
    \centering\small
    \vspace{-15pt}
    \caption{\textbf{Run time analysis} during training on ModelNet40 and S-PB\_T50\_RS variant of ScanObjectNN dataset. We report the time count in \textit{seconds}.}
    \label{tab:training_time}
    \begin{tabular}{l c c}
        \toprule
        \multirow{2}{*}{Method} & \multicolumn{2}{c}{\textbf{Dataset}} \\\cline{2-3}
         &   MN40 & S-PB\_T50\_RS \\
         \midrule
         {ULIP-2} &  132 & 106  \\
        {PointPRC} &  159 & 112  \\
         \rowcolor{skyblue} {\textbf{Refine3D}} & 162 & 115  \\
        \bottomrule
    \end{tabular}
    \vspace{-35pt}
\end{wraptable}

In Table \ref{tab:training_time}, we examine the training efficiency of our proposed method, ReFine3D. All methods are trained for 20 epochs using \textit{ NVIDIA V100 GPU}. On both datasets, ReFine3D shows a marginal increase in training time. This is due to tuning additional layers. However, the cost remains competitive and well within practical limits compared to the existing SOTA. Since ReFine3D does not add any extra parameters (e.g. prompts, adapters, like existing methods), the computational cost and memory footprint at inference remain the same as the pre-trained encoder.

\subsection{Scalability Analysis}
We further evaluate our framework on the larger and more challenging Objaverse-LVIS \citep{deitke2023objaverse} dataset, which serves as the target domain in this experiment. Objaverse-LVIS is a curated subset of the recently released Objaverse, containing 46,205 point clouds across 1,156 classes, including only a single instance for some classes, making it particularly difficult for conventional point cloud recognition models. As shown in Table~\ref{tab:scalability}, ReFine3D achieves substantial improvements, a +1.80\% gain on the source and a significant +3.36\% gain on the extensive target dataset.

\begin{table}
    \centering\small
    \vspace{-20pt}
    \caption{\textbf{Analysis on scalability}. Here, we use a smaller dataset ShapeNetV2, as the source domain and Objaverse-LVIS as the target domain. 
    }
    \label{tab:scalability}
    \begin{tabular}{l c c c}
        \toprule
        {\multirow{2}{*}{Method}} & \textbf{Source} & & \textbf{Target} \\\cline{2-4}
        & ShapeNetV2 & & Objaverse-LVIS \\
        \midrule
        ULIP-2 & 76.70(1.37) & & 14.80(0.22) \\
        PointPRC & 76.70(1.59) & & 18.07(0.49) \\
        \rowcolor{skyblue}
      \textbf{ReFine3D} & \textbf{78.50} (1.20) & & \textbf{21.43}(0.41) \\
     $\Delta$ & \improve{1.80} & & \improve{3.36} \\
        \bottomrule
    \end{tabular}
\end{table}

\subsection{Test-time Augmentation Trade-offs}
\begin{table}[t]
    \centering\small
    \caption{\textbf{Test-time augmentation trade-offs} on S-PB\_T50\_RS. We report 
    accuracy (HM), inference latency per sample, and throughput. All measurements 
    on NVIDIA V100 GPU. Here, $K_{test}$ represents number of augmented point clouds and $Q$ represents number of text prompts.}
    \label{tab:test_time_tradeoff}
    \begin{tabular}{l c c c c c}
        \toprule
        \textbf{Configuration} & \textbf{$K_{test}$} & \textbf{$Q$} & \textbf{HM} (\%) & \textbf{Latency (ms)} \\
        \midrule
        Single view (baseline) & 1 & 1 & 74.35 & 45 \\
        TTA (5 views) & 5 & 1 & 74.91 & 49 \\
        Text diversity & 1 & 7 & 74.68 & 46 \\
        \rowcolor{skyblue}TTA + text diversity (ours) & 5 & 7 & 75.90 & 50 \\
        \bottomrule
    \end{tabular}
\end{table}
Our test-time augmentation mechanism improves performance with minimal computational overhead. Table~\ref{tab:test_time_tradeoff} shows the accuracy-latency trade-off on S-PB\_T50\_RS using an NVIDIA V100 GPU. Single-view inference achieves 74.35\% HM with 45ms latency. Adding test-time augmentation ($K_{test}=5$) improves accuracy to 74.91\% (+0.56\%) with only 49ms latency. Text diversity ($Q=7$ prompts) provides 74.68\% accuracy with 46ms latency, as precomputed text embeddings add virtually no overhead. Our full configuration ($K_{test}=5$, $Q=7$) achieves 75.90\% HM with 50ms latency---a +1.55\% accuracy gain for only 5ms additional cost. The modest overhead stems from encoding augmented point clouds through the 3D encoder, while text diversity introduces negligible cost since embeddings are precomputed and cached.

\subsection{Discussion}
Our work provides several incremental insights that advance understanding of 3D multimodal fine-tuning:

\textbf{(1) Modality-Aware Regularization Synergy.} Table~\ref{tab:comprehensive_ablation} shows that augmentation-based regularization alone provides limited gains, but when combined with vision-guided multimodal alignment, contributions compound substantially to achieve 4.23\% total improvement. This demonstrates that integrating consistency regularization across point clouds with multimodal alignment is essential for learning generalizable representations under domain shift.

\textbf{(2) Quality vs. Quantity in Vision Supervision.} Table~\ref{tab:ablation_views} reveals that increasing rendered views paradoxically degrades novel class accuracy (from 75.80\% to 70.30\%), indicating view-specific overfitting. This counter-intuitive finding shows that strategic constraint in multimodal alignment---using minimal yet sufficient supervision---is more important than quantity for preserving generalization.

\textbf{(3) Asymmetric Encoder Tuning.} Table~\ref{tab:ablate_layers} shows layer-selective fine-tuning achieves 75.90\% HM versus 71.79\% for full fine-tuning, demonstrating that asymmetrically adapting only the 3D encoder while freezing shared semantic encoders prevents cross-modal drift and improves domain shift robustness. These highlight ReFine3D’s potential for scalable and generalizable adaptation of 3D foundation models.

\section{Conclusion}
In this paper, we propose ReFine3D, a regularized fine-tuning framework for 3D multi-modal foundation models. Our method tackles two key challenges: overlooking the unique characteristics of 3D point clouds and under-utilization of pre-trained VLMs' visual semantics. ReFine3D addresses these issues through selective layer tuning, consistency-based regularization, and test-time augmentation with confidence-based aggregation strategies. Furthermore, ReFine3D uses the pre-trained VLMs visual knowledge by aligning point cloud-text-vision triplets in a shared representation space. We demonstrate its effectiveness across multiple 3D domain generalization benchmarks, showing consistent gains in task-specific performance and cross-domain robustness.

\bibliography{tmlr}

@inproceedings{pointclip,
  title={Pointclip: Point cloud understanding by clip},
  author={Zhang, Renrui and Guo, Ziyu and Zhang, Wei and Li, Kunchang and Miao, Xupeng and Cui, Bin and Qiao, Yu and Gao, Peng and Li, Hongsheng},
  booktitle={Proceedings of the IEEE/CVF conference on computer vision and pattern recognition},
  pages={8552--8562},
  year={2022}
}

@inproceedings{pointclip2,
  title={Pointclip v2: Prompting clip and gpt for powerful 3d open-world learning},
  author={Zhu, Xiangyang and Zhang, Renrui and He, Bowei and Guo, Ziyu and Zeng, Ziyao and Qin, Zipeng and Zhang, Shanghang and Gao, Peng},
  booktitle={Proceedings of the IEEE/CVF International Conference on Computer Vision},
  pages={2639--2650},
  year={2023}
}

@inproceedings{clip2point,
  title={Clip2point: Transfer clip to point cloud classification with image-depth pre-training},
  author={Huang, Tianyu and Dong, Bowen and Yang, Yunhan and Huang, Xiaoshui and Lau, Rynson WH and Ouyang, Wanli and Zuo, Wangmeng},
  booktitle={Proceedings of the IEEE/CVF International Conference on Computer Vision},
  pages={22157--22167},
  year={2023}
}

@inproceedings{clip,
  title={Learning transferable visual models from natural language supervision},
  author={Radford, Alec and Kim, Jong Wook and Hallacy, Chris and Ramesh, Aditya and Goh, Gabriel and Agarwal, Sandhini and Sastry, Girish and Askell, Amanda and Mishkin, Pamela and Clark, Jack and others},
  booktitle={International conference on machine learning},
  pages={8748--8763},
  year={2021},
  organization={PMLR}
}

@inproceedings{wortsman2022robust,
  title={Robust fine-tuning of zero-shot models},
  author={Wortsman, Mitchell and Ilharco, Gabriel and Kim, Jong Wook and Li, Mike and Kornblith, Simon and Roelofs, Rebecca and Lopes, Raphael Gontijo and Hajishirzi, Hannaneh and Farhadi, Ali and Namkoong, Hongseok and others},
  booktitle={Proceedings of the IEEE/CVF conference on computer vision and pattern recognition},
  pages={7959--7971},
  year={2022}
}

@article{zhou2022domain,
  title={Domain generalization: A survey},
  author={Zhou, Kaiyang and Liu, Ziwei and Qiao, Yu and Xiang, Tao and Loy, Chen Change},
  journal={IEEE Transactions on Pattern Analysis and Machine Intelligence},
  volume={45},
  number={4},
  pages={4396--4415},
  year={2022},
  publisher={IEEE}
}

@inproceedings{modelnetC,
  title={Benchmarking and analyzing point cloud classification under corruptions},
  author={Ren, Jiawei and Pan, Liang and Liu, Ziwei},
  booktitle={International Conference on Machine Learning},
  pages={18559--18575},
  year={2022},
  organization={PMLR}
}

@article{pdg,
  title={Learning generalizable part-based feature representation for 3d point clouds},
  author={Wei, Xin and Gu, Xiang and Sun, Jian},
  journal={Advances in Neural Information Processing Systems},
  volume={35},
  pages={29305--29318},
  year={2022}
}

@article{pointdan,
  title={Pointdan: A multi-scale 3d domain adaption network for point cloud representation},
  author={Qin, Can and You, Haoxuan and Wang, Lichen and Kuo, C-C Jay and Fu, Yun},
  journal={Advances in Neural Information Processing Systems},
  volume={32},
  year={2019}
}

@inproceedings{metasets,
  title={Metasets: Meta-learning on point sets for generalizable representations},
  author={Huang, Chao and Cao, Zhangjie and Wang, Yunbo and Wang, Jianmin and Long, Mingsheng},
  booktitle={Proceedings of the IEEE/CVF Conference on Computer Vision and Pattern Recognition},
  pages={8863--8872},
  year={2021}
}

@inproceedings{ulip2,
  title={Ulip-2: Towards scalable multimodal pre-training for 3d understanding},
  author={Xue, Le and Yu, Ning and Zhang, Shu and Panagopoulou, Artemis and Li, Junnan and Mart{\'\i}n-Mart{\'\i}n, Roberto and Wu, Jiajun and Xiong, Caiming and Xu, Ran and Niebles, Juan Carlos and others},
  booktitle={Proceedings of the IEEE/CVF Conference on Computer Vision and Pattern Recognition},
  pages={27091--27101},
  year={2024}
}

@inproceedings{ulip,
  title={Ulip: Learning a unified representation of language, images, and point clouds for 3d understanding},
  author={Xue, Le and Gao, Mingfei and Xing, Chen and Mart{\'\i}n-Mart{\'\i}n, Roberto and Wu, Jiajun and Xiong, Caiming and Xu, Ran and Niebles, Juan Carlos and Savarese, Silvio},
  booktitle={Proceedings of the IEEE/CVF conference on computer vision and pattern recognition},
  pages={1179--1189},
  year={2023}
}

@inproceedings{idpt,
  title={Instance-aware dynamic prompt tuning for pre-trained point cloud models},
  author={Zha, Yaohua and Wang, Jinpeng and Dai, Tao and Chen, Bin and Wang, Zhi and Xia, Shu-Tao},
  booktitle={Proceedings of the IEEE/CVF International Conference on Computer Vision},
  pages={14161--14170},
  year={2023}
}

@inproceedings{ppt,
  title={Parameter-efficient prompt learning for 3d point cloud understanding},
  author={Sun, Hongyu and Wang, Yongcai and Chen, Wang and Deng, Haoran and Li, Deying},
  booktitle={2024 IEEE International Conference on Robotics and Automation (ICRA)},
  pages={9478--9486},
  year={2024},
  organization={IEEE}
}

@inproceedings{pointPEFT,
  title={Point-peft: Parameter-efficient fine-tuning for 3d pre-trained models},
  author={Tang, Yiwen and Zhang, Ray and Guo, Zoey and Ma, Xianzheng and Zhao, Bin and Wang, Zhigang and Wang, Dong and Li, Xuelong},
  booktitle={Proceedings of the AAAI Conference on Artificial Intelligence},
  volume={38},
  number={6},
  pages={5171--5179},
  year={2024}
}

@inproceedings{dept,
  title={Dynamic adapter meets prompt tuning: Parameter-efficient transfer learning for point cloud analysis},
  author={Zhou, Xin and Liang, Dingkang and Xu, Wei and Zhu, Xingkui and Xu, Yihan and Zou, Zhikang and Bai, Xiang},
  booktitle={Proceedings of the IEEE/CVF Conference on Computer Vision and Pattern Recognition},
  pages={14707--14717},
  year={2024}
}

@inproceedings{paul2023crossmoco,
  title={Crossmoco: multi-modal momentum contrastive learning for point cloud},
  author={Paul, Sneha and Patterson, Zachary and Bouguila, Nizar},
  booktitle={2023 20th Conference on Robots and Vision (CRV)},
  pages={273--280},
  year={2023},
  organization={IEEE}
}

@inproceedings{pointprc,
  title={Point-PRC: A Prompt Learning Based Regulation Framework for Generalizable Point Cloud Analysis},
  author={Sun, Hongyu and Ke, Qiuhong and Wang, Yongcai and Chen, Wang and Yang, Kang and Li, Deying and Cai, Jianfei},
  booktitle={The Thirty-eighth Annual Conference on Neural Information Processing Systems},
year={2024}
}

@inproceedings{pointnet,
  title={Pointnet: Deep learning on point sets for 3d classification and segmentation},
  author={Qi, Charles R and Su, Hao and Mo, Kaichun and Guibas, Leonidas J},
  booktitle={Proceedings of the IEEE conference on computer vision and pattern recognition},
  pages={652--660},
  year={2017}
}

@article{guo2020deep,
  title={Deep Learning for 3D Point Clouds: A Survey},
  author={Guo, Yulan and Wang, Hanyun and Hu, Qingyong and Liu, Hao and Liu, Li and Bennamoun, Mohammed},
  journal={IEEE Transactions on Pattern Analysis and Machine Intelligence},
  volume={43},
  number={12},
  pages={4338--4364},
  year={2021}
}

@article{qi2023fine,
  title={Fine-tuning aligned language models compromises safety, even when users do not intend to!},
  author={Qi, Xiangyu and Zeng, Yi and Xie, Tinghao and Chen, Pin-Yu and Jia, Ruoxi and Mittal, Prateek and Henderson, Peter},
  journal={arXiv preprint arXiv:2310.03693},
  year={2023}
}

@inproceedings{pointbert,
  title={Point-bert: Pre-training 3d point cloud transformers with masked point modeling},
  author={Yu, Xumin and Tang, Lulu and Rao, Yongming and Huang, Tiejun and Zhou, Jie and Lu, Jiwen},
  booktitle={Proceedings of the IEEE/CVF conference on computer vision and pattern recognition},
  pages={19313--19322},
  year={2022}
}

@inproceedings{promptsrc,
  title={Self-regulating prompts: Foundational model adaptation without forgetting},
  author={Khattak, Muhammad Uzair and Wasim, Syed Talal and Naseer, Muzammal and Khan, Salman and Yang, Ming-Hsuan and Khan, Fahad Shahbaz},
  booktitle={Proceedings of the IEEE/CVF international conference on computer vision},
  pages={15190--15200},
  year={2023}
}

@inproceedings{promptkd,
  title={Promptkd: Unsupervised prompt distillation for vision-language models},
  author={Li, Zheng and Li, Xiang and Fu, Xinyi and Zhang, Xin and Wang, Weiqiang and Chen, Shuo and Yang, Jian},
  booktitle={Proceedings of the IEEE/CVF Conference on Computer Vision and Pattern Recognition},
  pages={26617--26626},
  year={2024}
}

@article{shapenet,
  title={Shapenet: An information-rich 3d model repository},
  author={Chang, Angel X and Funkhouser, Thomas and Guibas, Leonidas and Hanrahan, Pat and Huang, Qixing and Li, Zimo and Savarese, Silvio and Savva, Manolis and Song, Shuran and Su, Hao and others},
  journal={arXiv preprint arXiv:1512.03012},
  year={2015}
}

@inproceedings{scanobject,
  title={Revisiting point cloud classification: A new benchmark dataset and classification model on real-world data},
  author={Uy, Mikaela Angelina and Pham, Quang-Hieu and Hua, Binh-Son and Nguyen, Thanh and Yeung, Sai-Kit},
  booktitle={Proceedings of the IEEE/CVF international conference on computer vision},
  pages={1588--1597},
  year={2019}
}

@inproceedings{modelnet,
  title={3d shapenets: A deep representation for volumetric shapes},
  author={Wu, Zhirong and Song, Shuran and Khosla, Aditya and Yu, Fisher and Zhang, Linguang and Tang, Xiaoou and Xiao, Jianxiong},
  booktitle={Proceedings of the IEEE conference on computer vision and pattern recognition},
  pages={1912--1920},
  year={2015}
}

@article{coop,
  title={Learning to prompt for vision-language models},
  author={Zhou, Kaiyang and Yang, Jingkang and Loy, Chen Change and Liu, Ziwei},
  journal={International Journal of Computer Vision},
  volume={130},
  number={9},
  pages={2337--2348},
  year={2022},
  publisher={Springer}
}

@inproceedings{zhang2023learning,
  title={Learning 3d representations from 2d pre-trained models via image-to-point masked autoencoders},
  author={Zhang, Renrui and Wang, Liuhui and Qiao, Yu and Gao, Peng and Li, Hongsheng},
  booktitle={Proceedings of the IEEE/CVF Conference on Computer Vision and Pattern Recognition},
  pages={21769--21780},
  year={2023}
}

@article{gao2024clip,
  title={Clip-adapter: Better vision-language models with feature adapters},
  author={Gao, Peng and Geng, Shijie and Zhang, Renrui and Ma, Teli and Fang, Rongyao and Zhang, Yongfeng and Li, Hongsheng and Qiao, Yu},
  journal={International Journal of Computer Vision},
  volume={132},
  number={2},
  pages={581--595},
  year={2024},
  publisher={Springer}
}

@article{song2023meta,
  title={Meta-adapter: An online few-shot learner for vision-language model},
  author={Song, Lin and Xue, Ruoyi and Wang, Hang and Sun, Hongbin and Ge, Yixiao and Shan, Ying and others},
  journal={Advances in Neural Information Processing Systems},
  volume={36},
  pages={55361--55374},
  year={2023}
}

@article{wei2022chain,
  title={Chain-of-thought prompting elicits reasoning in large language models},
  author={Wei, Jason and Wang, Xuezhi and Schuurmans, Dale and Bosma, Maarten and Xia, Fei and Chi, Ed and Le, Quoc V and Zhou, Denny and others},
  journal={Advances in neural information processing systems},
  volume={35},
  pages={24824--24837},
  year={2022}
}

@article{turpin2023language,
  title={Language models don't always say what they think: Unfaithful explanations in chain-of-thought prompting},
  author={Turpin, Miles and Michael, Julian and Perez, Ethan and Bowman, Samuel},
  journal={Advances in Neural Information Processing Systems},
  volume={36},
  pages={74952--74965},
  year={2023}
}

@inproceedings{deitke2023objaverse,
  title={Objaverse: A universe of annotated 3d objects},
  author={Deitke, Matt and Schwenk, Dustin and Salvador, Jordi and Weihs, Luca and Michel, Oscar and VanderBilt, Eli and Schmidt, Ludwig and Ehsani, Kiana and Kembhavi, Aniruddha and Farhadi, Ali},
  booktitle={Proceedings of the IEEE/CVF conference on computer vision and pattern recognition},
  pages={13142--13153},
  year={2023}
}

@article{miller1995wordnet,
  title={WordNet: a lexical database for English},
  author={Miller, George A},
  journal={Communications of the ACM},
  volume={38},
  number={11},
  pages={39--41},
  year={1995},
  publisher={ACM New York, NY, USA}
}

@inproceedings{zhang2024invariantoodg,
  title={InvariantOODG: Learning Invariant Features of Point Clouds for Out-of-Distribution Generalization},
  author={Zhang, Zhimin and Gao, Xiang and Hu, Wei},
  booktitle={ICASSP 2024-2024 IEEE International Conference on Acoustics, Speech and Signal Processing (ICASSP)},
  pages={7420--7424},
  year={2024},
  organization={IEEE}
}

@article{denninger2019blenderproc,
  title={Blenderproc},
  author={Denninger, Maximilian and Sundermeyer, Martin and Winkelbauer, Dominik and Zidan, Youssef and Olefir, Dmitry and Elbadrawy, Mohamad and Lodhi, Ahsan and Katam, Harinandan},
  journal={arXiv preprint arXiv:1911.01911},
  year={2019}
}

@article{oord2018representation,
  title={Representation learning with contrastive predictive coding},
  author={Oord, Aaron van den and Li, Yazhe and Vinyals, Oriol},
  journal={arXiv preprint arXiv:1807.03748},
  year={2018}
}

@inproceedings{uni3d,
  title={Uni3D: Exploring Unified 3D Representation at Scale},
  author={Zhou, Junsheng and Wang, Jinsheng and Ma, Baorui and Liu, Yu-Shen and Huang, Tiejun and Wang, Xinlong},
  booktitle={ICLR},
  year={2024}
}

@article{openshape,
  title={Openshape: Scaling up 3d shape representation towards open-world understanding},
  author={Liu, Minghua and Shi, Ruoxi and Kuang, Kaiming and Zhu, Yinhao and Li, Xuanlin and Han, Shizhong and Cai, Hong and Porikli, Fatih and Su, Hao},
  journal={Advances in neural information processing systems},
  volume={36},
  pages={44860--44879},
  year={2023}
}

@inproceedings{sun2025point,
  title={Point-Cache: Test-time Dynamic and Hierarchical Cache for Robust and Generalizable Point Cloud Analysis},
  author={Sun, Hongyu and Ke, Qiuhong and Cheng, Ming and Wang, Yongcai and Li, Deying and Gou, Chenhui and Cai, Jianfei},
  booktitle={Proceedings of the Computer Vision and Pattern Recognition Conference},
  pages={1263--1275},
  year={2025}
}

@article{hu2022lora,
  title={Lora: Low-rank adaptation of large language models.},
  author={Hu, Edward J and Shen, Yelong and Wallis, Phillip and Allen-Zhu, Zeyuan and Li, Yuanzhi and Wang, Shean and Wang, Liang and Chen, Weizhu and others},
  journal={Iclr},
  volume={1},
  number={2},
  pages={3},
  year={2022}
}

@article{paul2026adapter,
  title={An Adapter-free Fine-tuning Approach for Tuning 3D Foundation Models},
  author={Paul, Sneha and Patterson, Zachary and Bouguila, Nizar},
  journal={arXiv preprint arXiv:2603.23730},
  year={2026}
}

@inproceedings{paul2026point,
  title={Point Cloud as a Foreign Language for Multi-modal Large Language Model},
  author={Paul, Sneha and Patterson, Zachary and Bouguila, Nizar},
  booktitle={Proceedings of the IEEE/CVF Conference on Computer Vision and Pattern Recognition},
  pages={16676--16687},
  year={2026}
}

@inproceedings{paul2024improving,
  title={Improving 3D Semi-supervised Learning by Effectively Utilizing All Unlabelled Data},
  author={Paul, Sneha and Patterson, Zachary and Bouguila, Nizar},
  booktitle={European Conference on Computer Vision},
  pages={55--71},
  year={2024},
  organization={Springer}
}
\bibliographystyle{tmlr}

\end{document}